\documentclass[11pt]{article}

\usepackage{amsmath,amssymb,amsthm,bm,mathtools,graphicx,fullpage,natbib}
\usepackage[colorlinks,citecolor=blue]{hyperref}

\DeclareMathOperator*{\arginf}{arg\,inf}

\newtheorem{assumption}{Assumption}
\newtheorem{proposition}{Proposition}
\newtheorem{remark}{Remark}
\newtheorem{lemma}{Lemma}
\newtheorem{theorem}{Theorem}
\newtheorem{corollary}{Corollary}

\newcommand{\nimplies}{\mathrel{{\ooalign{\hidewidth$\not\phantom{=}$\hidewidth\cr$\implies$}}}}

\begin{document}

\title{Causal Learning via Manifold Regularization}

\author{Steven M. Hill$^{1,\ast}$ \and Chris J. Oates$^{2,\ast}$ \and Duncan A. Blythe$^3$ \and Sach Mukherjee$^{3,\ast}$ \vspace{0.1cm} \and
\small $^{1}$MRC Biostatistics Unit, University of Cambridge, Cambridge, UK.\\
\small $^{2}$School of Mathematics, Statistics and Physics, Newcastle University, Newcastle upon Tyne, UK.\\
\small $^{3}$German Centre for Neurodegenerative Diseases (DZNE), Bonn, Germany.\\
\small $^{\ast}$These authors contributed equally to this work.
			}

\maketitle

\begin{abstract}
This paper frames causal structure estimation as a machine learning task.
The idea is to treat indicators of causal relationships between variables as `labels' and to exploit available data on the variables of interest to provide features for the labelling task.
Background scientific knowledge or any available interventional data provide labels on some causal relationships and the remainder are treated as unlabelled. 
To illustrate the key ideas, we develop a distance-based approach (based on  bivariate histograms) within a manifold regularization framework. 
We present empirical results on three different  biological data sets
(including examples where  causal effects can be verified by experimental intervention), that together demonstrate the efficacy and general nature of the approach as well as its simplicity from a user's point of view.
\end{abstract}

\section{Introduction}
Causal structure learning is concerned with learning causal relationships between variables. 
Such relationships are often represented using directed graphs with nodes corresponding to  the variables of interest. 
Consider a set of $p$ variables or nodes indexed by $V = \{ 1 \ldots p \}$. 
The  aspect  we focus on in this paper is to  determine, for each (ordered) pair $(i,j) \in V {\times} V$, whether or not node $i$ exerts a causal influence on node $j$.  
In particular, our focus is on the binary `detection' problem (of learning whether or not node $i$ exerts a causal influence on node $j$) rather than estimation of the magnitude of any causal effect.

Methods for learning causal structures can be usefully classified according to whether the graph is intended to encode direct or total (ancestral) causal relationships.  
For example 
if variable $A$ acts on $B$ which in turn acts on $C$, $A$ has an ancestral effect on $C$ (via $B$). Here, the graph of direct effects has edges 
$A \rightarrow B \rightarrow C$, while the graph of total or ancestral effects has in addition the edge $A \rightarrow C$. 
Methods based on (causal) directed acyclic graphs (DAGs) are a natural and popular choice for causal discovery \citep{spirtes2000,pearl2009}.
The PC algorithm \citep{spirtes2000} is an important example of such a method. Using a sequence of tests of conditional independence, the PC algorithm estimates
an underlying causal DAG. Due to the fact that the graph may not be  identifiable, the output is an equivalence class of DAGs (encoded as a completed partially directed acyclic graph or CPDAG).
Here the estimand is intended to encode direct influences. 
IDA \citep[Intervention calculus when the DAG is Absent;][]{maathuis2009}
uses the PC output to bound the quantitative total causal effect of any node $i$ on any other node $j$. These estimated effects can be thresholded to provide a set of edges. 
FCI 
\citep[Fast Causal Inference;][]{spirtes2000}
and RFCI \citep[Really Fast Causal Inference;][]{colombo2012}  consider a type of ancestral graph  as estimand and  allow for latent variables. 
Greedy Interventional Equivalence Search \citep[GIES;][]{hauser2012} is a score-based approach that allows for the inclusion of interventional data.

Methods for learning causal structures (such as those above) are often rooted in data-generating causal models.
In a quite different vein,  there have been some interesting recent efforts in the direction of labelling pairs of variables as causal or otherwise, such as in \citet{lopez2015} and \citet{mooij2016}. 
These approaches are `discriminative' in spirit, in the sense that they need not be rooted in an explicit data-generating model; rather the emphasis is on learning how to tell causal and non-causal apart. 
Our work is in this latter vein.
We address a specific aspect of causal learning---that of estimating edges in a graph encoding causal relationships between a defined set of vertices---but via a machine learning approach that allows the inclusion of any available information concerning known cause-effect relationships.
The output of our method is a directed graph that need not be acyclic
\citep[see][for discussion of cyclic causality]{spirtes1995,richardson1996,hyttinen2012} 
 and whose edges may encode either direct or total/ancestral relationships, as discussed below.
 The main differences between our work and previous work on labelling causal pairs \citep{lopez2015,mooij2016} are the specific methods  and associated theory that we put forward, the manifold regularization framework,  and the empirical  examples.

In general terms the idea is as follows: 
let $\mathcal{D}$ denote the available data and $\Phi$ denote any available knowledge on causal relationships among the variables indexed in $V$ (e.g., based on background knowledge 
or experimental intervention).
We view the causal learning task in terms of constructing an estimator of the form $\hat{G}(\mathcal{D},\Phi)$,
where $\hat{G}$ is a directed graph with vertex set $V$ and edge set $E(\hat{G})$, with $(i,j) \! \in \! E(\hat{G})$ corresponding to the claim that variable $i$ has a causal influence on variable $j$. 
To put this another way: entries in a binary adjacency matrix encoding causal relationships  are treated as `labels' in a machine learning sense. 
From this point of view, the task 
of constructing the estimator $\hat{G}(\mathcal{D},\Phi)$
is essentially one of learning these labels from available data and from any {\it a priori} known labels (derived from $\Phi$). 
Thus, a key difference with respect to a number of existing methods is the nature of the inputs needed: our approach requires causal background information $\Phi$
as an input while several existing methods (such as PC) use only observational data.
The casual background information $\Phi$ need not be interventional data {\it per se}, but must encode knowledge on some causal relationships in the system (we consider both scenarios in empirical examples below).
Note also that 
in our approach 
the causal status of multiple pairs is coupled via the learning scheme: loosely speaking (see below for technical details), it is the position of a test pair on a classification manifold (relative to other pairs) that determines its status.

Our approach   differs in several ways from graphical model-based methods. In our approach, the same framework can be used to estimate either direct or ancestral causal relationships, depending on the precise input (we show real data examples of both tasks below). 
This is because the classifier can be agnostic to the label semantics: provided the Bayes' risk for the label of interest is sufficiently low, these labels can in principle be learned. 
In contrast to much of the  literature, our approach does not try to 
provide a full data-generating model of the causal system but instead focuses on the specific  problem of learning edges encoding causal relationships. As we see in experiments below, this can lead to good empirical performance, but the output is in a sense less rich than a full causal model (see the Discussion). Our work is motivated by scientific problems where 
good performance with respect to this narrower task can be useful in 
reducing  the hypothesis space and targeting future work.

The remainder of the paper is organized as follows. We first introduce some notation and discuss in more detail how causal learning can be viewed as a semi-supervised task. 
We then discuss a specific instantiation of the general approach, based on manifold regularization using a simple bivariate featurization.
Using this specific approach---which we call Manifold Regularized Causal Learning (MRCL)---we present empirical results using 
three biological data sets. The results cover a range of scenarios and include examples with explicitly interventional data.

\section{Methods}

\subsection{Notation}

Let $V =\{ 1, \ldots, p \}$ index a set of variables whose mutual causal relationships are of interest. 
Let $G$ denote a directed graph with vertex set $V$ and edge set $E$;
where useful, we use $V(G), E(G)$ to denote its vertex and edge sets and $A(G)$ to denote  the corresponding $p {\times} p $ binary adjacency matrix.
To make the connection between causal relationships and machine learning more transparent, we introduce linear indexing by $[k]$ of the pairs $(i,j)  \! \in \! V {\times} V$.
Where needed, we make the correspondence explicit, 
denoting by $(i[k],j[k])$ the variable pair corresponding to linear index $[k]$ and by $[k(i,j)]$ the linear index for pair $(i,j)$. 
Suppose $A$ is the adjacency matrix of the unknown graph of interest.
Let $y_{[k]} \!  \in \! \{-1,+1\}$ be a binary variable (for convenience mapped onto $\{-1,+1\}$) corresponding to the entry $(i[k],j[k])$ in $A$; 
these $y_{[k]}$'s are the labels or outputs to be learned.
Available data are denoted $\mathcal{D}$. 
Available {\it a priori} knowledge about causal relationships between the variables $V$ is denoted $\Phi$.

\subsection{Causal Semantics}
Given data $\mathcal{D}$ and background knowledge $\Phi$ we aim to construct an estimate $\hat{G}$, the latter being a directed graph that need not be acyclic. The information in $\Phi$ guides the learner.
Two main cases arise, both of which we consider in experiments below:
\begin{itemize}
\item {\it Total or ancestral effects}. 
Here, $\Phi$ contains information on total effects---for example via interventional experiments as  performed in  biology---and the edges
in the estimate  $\hat{G}$ are intended to describe such effects. This means that an edge $(i,j) \! \in \! E(\hat{G})$ is interpreted to mean that node $i$ is inferred to be a causal ancestor of node $j$. 
\item {\it Direct effects}. 
Here, $\Phi$  contains information on direct effects (relative to the variable set $V$) and the edges in the estimated graph $\hat{G}$ are intended to describe direct effects. Then, an edge $(i,j) \! \in \! E(\hat{G})$ is interpreted to mean that $i$ is inferred to be a direct cause of $j$ (relative to the variable set $V$). 
\end{itemize}

Our immediate motivation comes from the experimental sciences and we focus in particular on causal influences 
that can, at least in principle, be experimentally verified (even in the presence of latent variables) 
and where causal cycles are possible (as is often the case in biology or economics, see e.g., \citealp{hyttinen2012}). Accordingly, we do not demand acyclicity. 
In our empirical work in biology, the nature of the underlying chemical/physical system means that there are many 
small magnitude causal effects that are essentially irrelevant for the scientific context
and this is a characteristic of many  problem settings in the natural and social sciences.
This motivates a
pragmatic approach assuming that estimated graphs are not very dense or fully connected nor necessarily transitive\footnote{We emphasize that these are pragmatic assumptions motivated by the nature of experimental data and scientific applications, and not intended to be fundamental statements about causality. For example, \citet{hyttinen2012} make the point that cycles can be removed by considering time-varying data on a suitable time scale, but that nevertheless cycles are common in causal scientific models in economics, engineering and biology  due to the fact that measurements are usually taken at wider intervals.}.

\subsection{Semi-Supervised Causal Learning}
With the notation above, the task is to 
learn the $y_{[k]}$'s using $\mathcal{D}$ and $\Phi$.
This is done using a semi-supervised estimator $\hat{y}_{[k]}(\mathcal{D},\Phi)$ (we make the connection to semi-supervised learning explicit shortly).
For now assume availability of such an estimator (we discuss one specific approach below).
Then from the $\hat{y}_{[k]}$ we have an estimate of the graph of interest as $\hat{G}(\mathcal{D},\Phi)  =  (V, E(\hat{G}(\mathcal{D},\Phi)))$ (recall that the vertex set $V$ is known) with the edge set specified via the semi-supervised learner as
\begin{eqnarray}
(i,j) \in E(\hat{G}(\mathcal{D},\Phi)) & \iff & \hat{y}_{[k(i,j)]}(\mathcal{D},\Phi) = 1.
\label{G_from_Y}
\end{eqnarray}

Background  knowledge $\Phi$  could be based on relevant science or on available interventional data.
For example, in a given scientific setting, certain cause-effect information may be known from previous work or theory.
Alternatively, if some interventional data are available in the study at hand, this 
gives information on some causal relationships.
Whatever the source of the information, assume that it is known that certain pairs $(i,j)$ are either causal pairs (positive information) or not causal pairs (negative information). 
Using the notation above, this amounts to knowing, for some pairs $[k]$, the value of $y_{[k]}$.
In semi-supervised learning terms, the pairs whose causal status is known correspond to the labelled objects and the remaining pairs are the unlabelled objects. 

For each pair $[k]$, some of the data, or some transformation thereof will be used as predictors or inputs, denote these generically as $g_{[k]}(\mathcal{D})$.
That is, $g_{[k]}$ is a featurization of the data, with the featurization  specific to variables $(i[k], j[k])$.
Let $\mathcal{K}$ be the set of linear indices (i.e., $[k] \in \mathcal{K}$ is a variable pair), $\mathcal{L} \subset \mathcal{K}$ be the variable pairs with labels available (via $\Phi$) and $\mathcal{U} = \mathcal{K} \setminus \mathcal{L}$ be the set of unlabelled pairs. 
Let $\mathbf{y}^{\mathcal{L}}$ be a binary vector comprising the $m_{\mathcal{L}} = |\mathcal{L}|$ available labels and $\mathbf{y}^{\mathcal{U}}$ be an unknown binary vector of length $m_{\mathcal{U}} = |\mathcal{U}|$. The available labels are determined by the background information $\Phi$ and we can write $\mathbf{y}^{\mathcal{L}}(\Phi)$ to make this explicit. 
A semi-supervised learner gives estimates 
 for the unlabelled objects, given the data and available labels. That is, an estimate 
of the form $\hat{\mathbf{y}}^{\mathcal{U}} (g(\mathcal{D}),\mathbf{y}^{\mathcal{L}}(\Phi))$. 
With these in hand we have estimates for all labels and therefore for all edges 
via (\ref{G_from_Y}).

Formulated in this way, it is clear that essentially any combination of featurization  $g$ and semi-supervised learner could be used in this setting. Below, as a practical example, we explore graph-based manifold learning \citep[following][]{Belkin2006} combined with a simple bivariate featurization.

\subsection{A Bivariate Featurization}
For  distance-based  learning, we require a distance measure between objects (here, variable pairs) $[k],[k'] \in \mathcal{K}$.
The simplest  candidate distance between variable pairs $[k], [k']$ is based only on the  bivariate distribution for the variables comprising the pairs (we make this notion precise below). 
Proofs of propositions appearing  in this Section are provided in Appendix A.

\subsubsection{Distance between variable pairs}
Let $\mathbf{Z}$ denote the $p$-dimensional random variable whose $n$ realizations $\mathbf{z}^{(l)}$, $l = 1,\dots,n$, comprise the data set $\mathcal{D}$. 
Assume $\mathbf{Z} \in \mathcal{Z}_p =  [z_{\text{min}},z_{\text{max}}]^p$ and that $\mathcal{Z}_p$ is endowed with the Borel $\sigma$-algebra $\mathcal{B}_p = \mathcal{B}(\mathcal{Z}_p)$.
Let $\mathcal{P}$ be the set of all twice continuously differentiable probability density functions, generically denoted $\pi$, with respect to Lebesgue measure $\Lambda_2$ on $(\mathcal{Z}_2,\mathcal{B}_2)$.
Let $\Pi_{[k]}$ be the bivariate (marginal) distribution for components $i[k],j[k] \in V$ of $\mathbf{Z}$.

\begin{assumption}
Each $\Pi_{[k]}$ admits a density function $\pi_{[k]} \in \mathcal{P}$.
\end{assumption}

If available, the densities $\pi_{[k]},\pi_{[k']}$ could be used to define a distance between the pairs $[k],[k']$.
Let $d_{\mathcal{P}} : \mathcal{P} \times \mathcal{P} \rightarrow [0,\infty)$ denote a pseudo-metric\footnote{Recall that a pseudo-metric $d$ satisfies all of the properties of a metric with the exception that $d(x,y) = 0 \nimplies x=y$.} on $\mathcal{P}$. Since we do not have access to the underlying probability density functions, we construct an analogue using the available data $\mathcal{D}$.
Let $\mathcal{S}_n := [z_{\text{min}},z_{\text{max}}]^{2n}$ denote the space of possible bivariate samples (the sample size is $n$) and $S_{[k]} \in \mathcal{S}_n$ denote the subset of the data for the variable pair $[k]$. That is,
$S_{[k]}  = \{ (z_{i[k]}^{(l)} , z_{j[k]}^{(l)}) \}_{l=1, \ldots, n} \subset \mathcal{Z}_2$. 

Let $\kappa : \mathcal{S}_n \rightarrow \mathcal{P}$ be a density estimator (DE).
We consider sample quantities of the form 
$d_{\mathcal{S}} = d_{\mathcal{P}} \circ (\kappa \times \kappa)$. That is, given data $S_{[k]}, S_{[k']} \in \mathcal{S}_n$ on two pairs $[k],[k']$, the DE is applied separately  to produce density estimates $\kappa(S_{[k]})$ and $\kappa(S_{[k']})$, 
that are compared using $d_{\mathcal{P}}$ to give  $d_{\mathcal{S}}(S_{[k]},S_{[k']}) = d_{\mathcal{P}}(\kappa(S_{[k]}) , \kappa(S_{[k']}))$.
This construction ensures that $d_{\mathcal{S}}$ is a pseudo-metric without  assumptions on the DE $\kappa$:

\begin{proposition} \label{prop: pseudometric inherit}
Assume that $d_{\mathcal{P}}$ is a pseudo-metric on $\mathcal{P}$.
Then $d_{\mathcal{S}}$ is a pseudo-metric on $\mathcal{S}_n$.
If, in addition, $\kappa$ is injective and $d_{\mathcal{P}}$ is a metric on $\mathcal{P}$, then $d_{\mathcal{S}}$ is a metric on $\mathcal{S}_n$.
\end{proposition}

\subsubsection{Choice of distance}

For semi-supervised learning we need a notion of distance under which causal pairs are relatively `close' to each other. 
For a measurable space $\mathcal{X}$ equipped with a measure $\rho$ we let $\|f\|_{L^q(\rho)} := \left( \int_{\mathcal{X}} |f|^q \mathrm{d}\rho \right)^{\frac{1}{q}} < \infty$.
The notion of distance that we consider is
$$
d_{\mathcal{P}}(\pi,\tilde{\pi}) := \|\pi - \tilde{\pi}\|_{L^2(\Lambda_2)}  .
$$
The right hand side exists since the integrand is continuous on a compact set and thus bounded.
This can be contrasted with the kernel embedding that was proposed for supervised causal learning  in \cite{lopez2015}.

\begin{proposition} \label{prop: d metric}
$d_{\mathcal{P}}$ is a metric on $\mathcal{P}$.
\end{proposition}

The main requirement that we have of the DE is that it provides consistent estimation in the $\|\cdot\|_{L^2(\Lambda_2)}$ norm when $\pi \in \mathcal{P}$.
Specifically, consider a sequence $S^{(n)}$ in $\mathcal{S}_n$ indexed by the number $n$ of data points.
In particular, suppose that $S^{(n)}$ is built from $n$ independent data points whose distribution is $\Pi$ (the shorthand notation $S^{(n)} \stackrel{\text{i.i.d.}}{\sim} \Pi$ will be used).
Let $\pi$ be the density function for $\Pi$.
Then $\kappa$ is said to be ``consistent'' if $\|\pi - \kappa(S^{(n)})\|_{L^2(\Lambda_2)} = o_P(1)$ holds for $S^{(n)} \stackrel{\text{i.i.d.}}{\sim} \Pi$ whenever $\pi \in \mathcal{P}$.

\begin{proposition} \label{prop: consistent}
Suppose $\kappa$ is consistent and that $\Pi, \tilde{\Pi}$ admit densities $\pi,\tilde{\pi} \in \mathcal{P}$.
Then, for $S^{(n)} \stackrel{\text{\emph{i.i.d.}}}{\sim} \Pi$, $\tilde{S}^{(n)} \stackrel{\text{\emph{i.i.d.}}}{\sim} \tilde{\Pi}$, where $S^{(n)}$ and $\tilde{S}^{(n)}$ are not necessarily independent, we have that $d_{\mathcal{S}}(S^{(n)},\tilde{S}^{(n)}) = d_{\mathcal{P}}(\pi,\tilde{\pi}) + o_P(1)$.
\end{proposition}
\noindent Thus $d_{\mathcal{S}}$ approximates the idealized metric $d_{\mathcal{P}}$ in the limit of  draws from $\Pi$ and $\tilde{\Pi}$.
Note that, in our intended use case, the $S^{(n)}$ and $\tilde{S}^{(n)}$ will correspond to bivariate scatter plots $S_{[k]}$, $S_{[k']}$ generated from the same underlying $\mathbf{z}^{(l)}$, $l = 1,\dots,n$, and hence $S^{(n)}$ and $\tilde{S}^{(n)}$ will not be independent.

For the experiments in this paper, motivated by computational ease, we used a simple bivariate histogram as the DE $\kappa$. To this end, partition $\mathcal{Z}_2$ into an $M \times M$ regular grid whose $(m_1,m_2)$th element is denoted $B_{m_1,m_2}$.
The standard bandwidth notation $h = M^{-1}$ will also be used.
For a scatter plot $S \in \mathcal{S}_n$, let $x_{m_1,m_2}$ denote the number of elements that belong to the set $B_{m_1,m_2}$.
Then the histogram estimator is
\begin{align}
\kappa(S)(\mathbf{z}') = \sum_{m_1,m_2=1}^M \frac{x_{m_1,m_2}}{n} \frac{1}{h^2} \mathbb{I}[\mathbf{z}' \in B_{m_1,m_2}] , \qquad \mathbf{z}' \in \mathcal{Z}_2. \label{eq: histogram estimator defn}
\end{align}
This DE is consistent in the sense of Proposition~\ref{prop: consistent}.
Indeed:

\begin{proposition} \label{prop: convergence rate}
Let the bandwidth parameter $h$ of the histogram estimator $\kappa$ be chosen such that $n h^2 \rightarrow \infty$.
Then $\kappa$ is consistent.
Moreover, an optimal choice of $h \asymp n^{-1/4}$ leads to $\|\pi - \kappa(S^{(n)})\|_{L^2(\Lambda_2)} = O_P(n^{-1/4})$ whenever $S^{(n)} \stackrel{\text{\emph{i.i.d.}}}{\sim} \Pi$ and $\pi \in \mathcal{P}$.
\end{proposition}

We note that this histogram DE is not rate optimal for the class $\mathcal{P}$ \citep[for comparison, kernel DEs attain a rate of $O_P(n^{-2/3})$ over the same class $\mathcal{P}$ of twice continuously differentiable bivariate densities considered here, see][]{wand1994}.
However, an important advantage of the histogram DE is that the subsequent evaluation of $\kappa(S)$ is $O(1)$, compared with $O(n)$ for the kernel DE.

\subsubsection{Implementation of the DE}
The above arguments support the use of a bivariate histogram to provide a simple featurization for variable pairs. 
In practice, for all examples below, the data were standardized, then truncated to $[-3,3]^2$, following which a bivariate histogram with bins of fixed width 0.2 was used.
The dimension of the resulting feature matrix was then reduced (to 100) using PCA.

\subsection{Manifold Regularization}

Recall that the goal is to estimate binary labels $\mathbf{y}^{\mathcal{U}}$ for a subset $\mathcal{U} \subset \mathcal{K}$ of variable pairs  given available data $\mathcal{D}$ and known labels $\mathbf{y}^{\mathcal{L}}(\Phi)$ for a subset $\mathcal{L} = \mathcal{K} \setminus \mathcal{U}$ (these are taken to be obtained from available interventional experiments and/or background knowledge). For any two pairs $[k],[k'] \in \mathcal{K}$, we also have available a distance $d_{\mathcal{S}}(S_{[k]}, S_{[k']})$. This is a  task in semi-supervised learning \citep[see e.g.][]{Belkin2006,Fergus2009} and a number of  formulations and methods could be used for estimation in this setting. Here we describe a specific approach in detail, using  manifold regularization methods discussed in \cite{Belkin2006}.

Let $\mathbf{x}_{[k]}$ denote a vector whose entries are the bin-counts $x_{i,j}$, $1 \leq i , j \leq M$, appearing in \eqref{eq: histogram estimator defn}, for scatter plot $S_{[k]}$.
Let $\mathcal{X} = \bigtimes_{1 \leq i,j \leq M} [0,n]$ and note that $\mathbf{x}_{[k]} \in \mathcal{X}$.
Then we make the observation that, for the histogram estimator,
$$
d_{\mathcal{S}}(S_{[k]} , S_{[k']}) \propto \| \mathbf{x}_{[k]} - \mathbf{x}_{[k']}\|_2 \, . 
$$
This perspective emphasizes that $g_{[k]}(\mathcal{D}) = \mathbf{x}_{[k]}$ is the featurization that underpins this work, and that the classification task can be considered as the construction of a map $c : \mathcal{X} \rightarrow \{-1,+1\}$.
To develop an approach to semi-supervised classification in the manner of \cite{Belkin2006}, let $\rho_{\mathcal{X}}$ be a reference measure on $\mathcal{X}$ and let $K : \mathcal{X} \times \mathcal{X} \rightarrow \mathbb{R}$ be a Mercer kernel; i.e., continuous, symmetric and positive semi-definite.
The reproducing kernel Hilbert space, $\mathcal{H}_K$, associated to $K$ can be defined via the integral operator $\Sigma_K : L^2(\rho_{\mathcal{X}}) \rightarrow L^2(\rho_{\mathcal{X}})$ where
$$
\Sigma_K f (\mathbf{x}) = \int K(\mathbf{x},\tilde{\mathbf{x}}) f(\tilde{\mathbf{x}}) \mathrm{d}\rho_X(\tilde{\mathbf{x}}).
$$
From the fact that $K$ is a Mercer kernel it follows that $\Sigma_K$ is self-adjoint, positive semi-definite and compact.
In particular, $\Sigma_K^\alpha$ is well-defined for $\alpha \in (0,\infty)$.
The reproducing kernel Hilbert space is defined as $\mathcal{H}_K = \Sigma_K^{\frac{1}{2}} L^2(\rho_{\mathcal{X}})$ and its norm is $\| f \|_{\mathcal{H}_K} := \| \Sigma_K^{- \frac{1}{2}} f \|_{L^2(\rho_{\mathcal{X}})}$; c.f. Corollary 4.13 in \cite{Cucker2007}.

Recall that $m_{\mathcal{L}} = |\mathcal{L}|$ is the number of available labels and $m_{\mathcal{U}} = |\mathcal{U}|$ the number of unlabelled pairs. Let $m=m_{\mathcal{U}} + m_{\mathcal{L}}$ ($=|\mathcal{K}|$) be the total number of pairs.
Using the distance function $d_{\mathcal{S}}$ we first define an $m \times m$ similarity matrix $\mathbf{W}$ with entries
\begin{equation}
W_{[k],[k']} = \exp\left(- \frac{1}{2\sigma_1^2} \| \mathbf{x}_{[k]} - \mathbf{x}_{[k']}\|_2^2 \right) \label{eq: similarity matrix}
\end{equation}
where $\sigma_1 > 0$ must be specified.
The squared-exponential form is motivated by an analytic connection between the heat kernel and the Laplace-Beltrami operator, which will be exploited in Section~\ref{subsubsec: belkin theory}.
We will use a partition of the matrix corresponding to the sets $\mathcal{U},\mathcal{L}$ as follows
$$
\mathbf{W} = \left[ \begin{array}{cc} \mathbf{W}^{\mathcal{LL}} & \mathbf{W}^{\mathcal{LU}} \\ \mathbf{W}^{\mathcal{UL}} & \mathbf{W}^{\mathcal{UU}} \end{array} \right]
$$

\noindent 
where we have assumed, without loss of generality, that the variable pairs are ordered so that the labelled pairs appear in the first $m_{\mathcal{L}}$ places, followed by the $m_{\mathcal{U}} = m - m_{\mathcal{L}}$ unlabelled pairs.
Correspondingly let 
$$
\mathbf{y} = \left[\begin{array}{c} \mathbf{y}^{\mathcal{L}} \\ \mathbf{y}^{\mathcal{U}} \end{array} \right] \in \{-1,+1\}^m
$$ 
denote a label matrix, where $+1$ indicate those pairs $[k]$ for which $y_{[k]}=1$. 
The vector $\mathbf{y}^{\mathcal{U}}$ is unknown and is the object of estimation.

Let $\mathbf{D}$ be the $m \times m$ diagonal matrix with diagonal entries $D_{[k],[k]} = \sum_{[k'] \in \mathcal{K}} W_{[k],[k']}$.
Define $\mathbf{L} = \mathbf{D} - \mathbf{W}$ (i.e., the un-normalized graph Laplacian; all matrices with $O(m^2)$ entries are denoted as bold capitals to emphasize the potential bottleneck that is associated with storage and manipulation of these matrices).
Let 
$$
\mathbf{f} = \left[ \begin{array}{c} \mathbf{f}^{\mathcal{L}} \\ \mathbf{f}^{\mathcal{U}} \end{array} \right] \in \mathbb{R}^m
$$ 
be a vector corresponding to a classification function $f : \mathcal{X} \rightarrow \mathbb{R}$ evaluated at the $m$ variable pairs $\mathcal{K}$, with the superscripts indicating correspondence with the labelled and unlabelled pairs.
Intuitively, we want the sign of $\bf{f}$ to agree with the known labels $\mathbf{y}^{\mathcal{L}}$ and also to take account of the manifold structure encoded in $\mathbf{L}$. 

In this work we consider a classifier of the form $\hat{c}(\mathbf{x}) = \text{sign}(\hat{f}(\mathbf{x}))$ where $\hat{f}$ arises from the \emph{Laplacian-regularized least squares} method
\begin{align}
\hat{f} = \arginf_{f \in \mathcal{H}_K } \frac{\| \mathbf{y}^{\mathcal{L}} - \mathbf{f}^{\mathcal{L}} \|_2^2 }{m_L} + \lambda_1 \frac{\mathbf{f}^\top \mathbf{L} \mathbf{f}}{m} + \lambda_2 \|f\|_{\mathcal{H}_K}^2 \, , \label{eq: LRLS}
\end{align}
following Section 4.2 of \cite{Belkin2006}.
Here the first term relates the known labels to the values of the function $f$. 
The second term imposes  `smoothness' on the label assignment 
in the sense of encouraging solutions where the labels do not change quickly with respect to the distance metric.
The third term is principally to ensure that the infimum remains well-defined and unique in the situation where there is insufficient data for the first penalty alone to be sufficient \citep[see Remark 2 in][]{Belkin2006}.

\begin{remark}[Choice of loss] \label{rem: loss function}
It is important to comment on our choice of a squared-error loss function in \eqref{eq: LRLS}, which differs from the more natural approach of using hinge loss for a binary classification task.
Our motivation here is principally computational expedience; the computational burden associated with the $m = O(p^2)$ different scatter plots requires that a light-weight estimation procedure is used.
However, we note that we are not the first to propose the use of squared-error loss in the classification context; it is in fact a standard approach to classification in the situation when the number of classes is $>2$ \citep[e.g.,][]{Wang2008}.
\end{remark}

\subsubsection{Consistency of the Classifier} \label{subsubsec: belkin theory}

As explained in Remark~\ref{rem: loss function}, the use of a squared-error loss function in a classification context is somewhat unnatural.
It is therefore incumbent on us to establish consistency of the proposed method.

To this end, we exploit the specific form of the similarity matrix used in \eqref{eq: similarity matrix}.
Indeed, if we re-write
\begin{align}
\frac{\mathbf{f}^\top \mathbf{L} \mathbf{f}}{m} = \frac{1}{2m} \sum_{[k] , [k'] \in \mathcal{K} } ( f(\mathbf{x}_{[k]}) - f(\mathbf{x}_{[k']}) )^2 W_{[k],[k']} \label{eq: manifold penalty}
\end{align}
then it can be established (under certain regularity conditions) that, if input data $\mathbf{x}$ are independently drawn from $\rho_\mathcal{X}$, then \eqref{eq: manifold penalty} converges to the quantity $\int f(\mathbf{x}) \Delta_{\mathcal{M}} f(\mathbf{x}) \mathrm{d}\rho_\mathcal{X}^2(\mathbf{x})$ (up to proportionality), a smoothness penalty based on weighted Laplace-Beltrami operator $\Delta_{\mathcal{M}}$ on the manifold $\mathcal{M}$ induced by $\rho_\mathcal{X}$ \citep{Grigoryan2006}.
The convergence occurs as $m, \sigma_1^2 m^{d+2} \rightarrow \infty$ \citep[Theorem 3.1 of][]{Belkin2008}.

This convergence of the graph Laplacian to the Laplace-Beltrami operator underlies existing consistency results for semi-supervised regression \citep[e.g.,][]{Cao2012} and is exploited again to establish the consistency of our classifier $\hat{c}(\mathbf{x}) = \text{sign}(\hat{f}(\mathbf{x}))$ in Appendix~\ref{app: convergence appendix}.
In summary, the ability to assign the correct label to an unlabelled pair $[k] \in \mathcal{L}$ depends on both the intrinsic predictability of the label as a function of the scatter plot $S_{[k]}$, as quantified by the Bayes risk, and the smoothness of the Bayes classifier $f_\rho$ as quantified by the largest value $\alpha \in (0,1]$ such that $\Sigma_K^{-\frac{\alpha}{2}} f_\rho \in L^2(\rho_{\mathcal{X}})$; see Corollary~\ref{cor: risk bound 2} in Appendix~\ref{app: convergence appendix} for full detail.

\subsubsection{Implementation of the Classifier}

Given training labels $\mathbf{y}^{\mathcal{L}}$, label estimates $\hat{\mathbf{y}}^{\mathcal{U}} = \text{sign}(\hat{\mathbf{f}}^{\mathcal{U}})$ are obtained by minimizing the objective function described above, as explained in Equation 8 in \cite{Belkin2006}. 
This gives
\begin{equation}
\hat{\mathbf{f}}^{\mathcal{U}} = \mathbf{K}_{\mathcal{U},\mathcal{K}} \left( \left[ \begin{array}{cc} \mathbf{I}_{m_{\mathcal{L}}} & \mathbf{0} \\ \mathbf{0} & \mathbf{0} \end{array} \right] \mathbf{K}_{\mathcal{K},\mathcal{K}} + \lambda_2 m_{\mathcal{L}} \mathbf{I}_m + \frac{\lambda_1 m_{\mathcal{L}}}{m^2} \mathbf{L} \mathbf{K}_{\mathcal{K},\mathcal{K}} \right)^{-1} \left[ \begin{array}{c} \mathbf{y}^{\mathcal{L}} \\ \mathbf{0} \end{array} \right]
\label{minimisation}
\end{equation}
where $\mathbf{K}_{\mathcal{U},\mathcal{K}}$ is the $m_{\mathcal{U}} \times m$ kernel matrix based on the unlabeled $\mathcal{U}$ and total $\mathcal{K}$ data, $\mathbf{K}_{\mathcal{K},\mathcal{K}}$ is the $m \times m$ kernel matrix based on the total data $\mathcal{K}$ and $\mathbf{I}_m$ denotes an $m$-dimensional identity matrix.

Here $\hat{\mathbf{y}}^U$
provides a point estimate for the unknown labels while $\hat{\mathbf{f}}^{\mathcal{U}}$ is real-valued and 
can be used to rank candidate pairs if required.
The linear system in (\ref{minimisation}) can be solved at a naive computational cost of $O(m^3)$. 
Computation for large-scale semi-supervised learning has  been studied in the literature \cite[see e.g.,][]{Fergus2009} and a number of approaches could be used to scale up to larger problems, but were not pursued in this work.

For experiments reported below 
we employed a similarity matrix (with length scale $\sigma_1$ as in 
\eqref{eq: similarity matrix})
and a kernel
$$
K(\mathbf{x},\mathbf{x}') =  \exp\left(- \frac{1}{2\sigma_2^2} \| \mathbf{x} - \mathbf{x}' \|_2^2 \right)
$$
whose length-scale parameter $\sigma_2$ was set  equal to $\sigma_1$ in the absence of prior knowledge about the manifold $\mathcal{M}$.
The scale $\sigma_1$ was set to the average distance to the nearest 50 points in the feature space (in practice estimated via a subsample).

The two penalty parameters in \eqref{eq: LRLS} were set to small positive values ($\lambda_1 = \lambda_2 = 0.001$; we found results were broadly insensitive to this choice). 
Following common practice we worked with the normalized graph Laplacian $\tilde{\mathbf{L}} := \mathbf{D}^{- \frac{1}{2}} \mathbf{L} \mathbf{D}^{- \frac{1}{2}}$ in place of $\mathbf{L}$ \citep[see Remark 3 of][]{Belkin2006}.

\section{Empirical Results} \label{sec: experimental results}

We tested our approach using three data sets with different characteristics. The key features of each data set are outlined below, with a full description of each data set appearing in the respective subsection. In all cases performance was assessed using either held-out interventional data or scientific knowledge.
\begin{itemize}
\item {\bf D1: Yeast knockout data.} Here, we used a data set due to \cite{kemmeren2014}, previously considered for causal learning in \cite{Peters2015,meinshausen2016}. The data consist of a large number of gene deletion experiments with corresponding gene expression measurements. 
\item {\bf D2: Kinase intervention data from human cancer cell lines.} These data, due to \cite{hill2017}, involve a small number of interventions on human cells, with corresponding protein measurements over time. 
\item {\bf D3: Protein data from cancer patient samples.}  These data arise from The Cancer Genome Atlas (TCGA) and are presented in \cite{akbani2014}. There are no interventional data, but the data pertain to relatively well-understood biological processes allowing  inferences to be checked against causal scientific knowledge.
\end{itemize}

An appealing feature of MRCL is the simplicity with which it can be applied to diverse problems. In each case below, we simply concatenate available data to form the data set $\mathcal{D}$ and available knowledge/interventions to form $\Phi$, then directly apply the methods as described.

\subsection{General Problem Set-Up}

The basic idea in all three problems was as follows: given data on a set of variables, for each (ordered) pair $(i,j)$ of variables we sought to determine  whether or not $i$ has a causal effect on $j$. 
In the case of data sets {\bf D1} and {\bf D2} the results were  assessed against the outcome of experiments involving explicit interventions. 
As discussed above, such experiments reveal ancestral relationships (that need not be direct)
and the goal in these examples was to learn such relationships.
The availability of a large number of interventions in {\bf D1} allowed a wider range of experiments, whereas {\bf D2} is a  smaller data set (but from human cells), allowing only a relatively limited assessment. In the case of {\bf D3}, where  interventional data (i.e. interventions on the same biological material that give rise to the training data) were not available but the relevant biological mechanisms are relatively well understood, we compared results to a reference mechanistic  graph derived from the domain literature. The literature itself is in effect an encoding of extensive interventional experiments combined with biochemical and biophysical knowledge. This gives information on direct edges and here the edges learned are intended to represent direct causes (relative to the  set of observed variables).
Within the semi-supervised set-up, a subset of pairs were labelled at the outset and the remaining pairs were unlabelled. 
All empirical results below are for unlabelled pairs; that is, in all cases assessment is carried out with respect to causal (and non-causal) relationships that were not used to train the models. 

\subsection{Data Set D1: Yeast Gene Expression}

\noindent
{\bf Data.} The data consisted of gene expression levels (log ratios) for a total of $p_{\mathrm{total}}=6170$ genes. 
Some of the data samples were measurements after knocking out a specific gene (interventional data) and the other samples were without any such intervention (observational data), with sample sizes of $n^{\mathrm{int}}=1479$ and $n^{\mathrm{obs}}=153$ respectively.
Each of the genes intervened on was one of the $p_{\mathrm{total}}$ genes. Let $t(l)$ be the index of the gene targeted by the $l^{\mathrm{th}}$ intervention. That is, the $l^{\mathrm{th}}$ interventional sample was an experiment in which gene $t(l)$ was knocked out. Let $T = \{ t(1), \ldots, t(n^{\mathrm{int}}) \}$ be the subset of genes that were the target of an interventional experiment.

\medskip
\noindent
{\bf Problem set-up.} Our problem set-up was as follows. We sampled a subset $C \subset T$ of the genes that were intervened upon, with  $|C|=50$, and treated this as the vertex set of interest (i.e., setting $V=C$ and $p=|C|=50$).
The goal was to uncover causal relationships between these $p$ variables. 

Since by design interventional data were available for all variables $j \! \in \! C$, we used these data to define an interventional `gold standard'. 
To this end we used a robust $z$-score that considered the change in a variable of interest under intervention, relative to its observational variation.
Let $Z_{ij}^{\mathrm{int}}$ denote the expression level of gene $j$ following intervention on gene $i$.
For any pair of genes $i,j \! \in \! C$ we say that gene $i$ has a causal effect on gene $j$ if and only if 
$\zeta_{ij}  =  |Z_{ij}^{\mathrm{int}} - M_j^{\mathrm{obs}}|/\text{IQR}_j^{\mathrm{obs}} \, > \tau$,
where $M_j^{\mathrm{obs}}$ is the median level of gene $j$ (calculated using half of the observational data samples; the remaining samples were used as training data---see below), 
$\text{IQR}_j^{\mathrm{obs}}$ the corresponding inter-quartile range and $\tau = 5$ was a fixed threshold. 
That is, we say there is an (experimentally verified) causal relationship between gene $i$ and gene $j$ if and only if $\zeta_{ij} \! > \! \tau$.
An absence of causal effects precludes estimation of true positive rates; hence we sampled $C$ subject to 
a sparsity condition (that at least 2.5\% of gene pairs show an effect).

Let $A(C)$ be a $p \times p$ binary matrix encoding the causal effects as described in the foregoing (i.e., $A(C)_{ij} = 1$ indicates that $i$ has an experimentally verified causal effect on $j$). Then, given data on genes $C$, we set up the learning problem as follows.
We treated a fraction $\rho$ of the entries in $A(C)$ as the available labels $\Phi$.
Thus, here $m = p^2 = 2500$, $m_{\mathcal{L}} = \lfloor \rho \,  m \rfloor$ and $m_{\mathcal{U}} = m-m_{\mathcal{L}}$.
Using these labels and data on the variables $C$, we  learned causal edges as described.
This gave estimates for the remaining (unseen) entries in $A(C)$, which we compared against the corresponding true values.  
The data set $\mathcal{D}$ comprised expression measurements for the genes in $C$ for $n_{\text{train}}^{\text{obs}}=76$ observational data samples (those samples not used to calculate the robust $z$-scores),  plus $n_{\text{train}}^{\text{int}}$ interventional data samples where genes outside the set of interest were intervened upon; that is, a subset of the 1429 genes in $T \setminus C$. 
This set-up ensured that $\mathcal{D}$ include neither any of the interventional nor observational data that was used to obtain the ground-truth matrix $A(C)$.
The total amount of training data is denoted by $n_{\text{train}} = n_{\text{train}}^{\text{obs}} + n_{\text{train}}^{\text{int}}$.
We considered $n_{\text{train}}=200, 500$ and $1000$ (corresponding to  $n_{\text{train}}^{\text{int}}=124, 424$ and $924$ respectively, sampled at random).

\medskip
\noindent
{\bf Results.} 
We compared the proposed Manifold Regularized Causal Learning  (MRCL) approach with the following approaches: 
\begin{itemize}
\item Penalized regression with an $\ell_1$ penalty \citep[{\bf Lasso};][]{Tibshirani1996}.
Each variable $j\in C$ was regressed on all other variables $i\in C, i\neq j$ to obtain regression coefficients.
This is not a causal approach as such, but is included as a simple multivariate baseline.

\item Intervention-calculus when the DAG is absent \citep[{\bf IDA};][]{maathuis2009,maathuis2010}.
A lower bound for the total causal effect of variable $i$ on variable $j$ was estimated for each pair $i,j\in C, i\neq j$.

\item The PC algorithm \citep[{\bf PC};][]{spirtes2000}. This provides a CPDAG estimate for the variables $C$. 

\item GIES \citep[{\bf GIES};][]{hauser2012}. This provides an essential graph estimate for the variables $C$, and allows inclusion of interventional data in a principled manner.

\end{itemize}

As simple baselines, we also included Pearson and Kendall correlation coefficients ({\bf Pearson} and {\bf Kendall})
and, following a suggestion from a referee, a simple k-nearest neighbor approach based on the featurization introduced above ({\bf k-NN}).

We note  that the causal methods compared against here differ in various ways from MRCL in the nature of their inputs and outputs and should not be regarded as direct competitors. Rather, the aim of the experiments is to investigate how MRCL performs on real data, whilst providing a set of baselines corresponding to well-known causal tools and standard correlation measures.

For the methods resulting in a score $s_{ij}$ for all pairs $i,j \in C, i\neq j$ (i.e., correlation or regression coefficients, total causal effects, or, for MRCL, the real-valued $\hat{\mathbf{f}}$ in (\ref{minimisation})), the scores were thresholded and pairs $(i,j)$ whose absolute values of the score fell above the threshold were labelled as `causal'. Varying the threshold and calculating true positives and false positives with respect to the binary \emph{unseen} entries in the matrix $A(C)$ resulted in a receiver operating characteristic (ROC) curve. 

\begin{figure}[tbh]
\centering
\includegraphics[width = 0.6\textwidth]{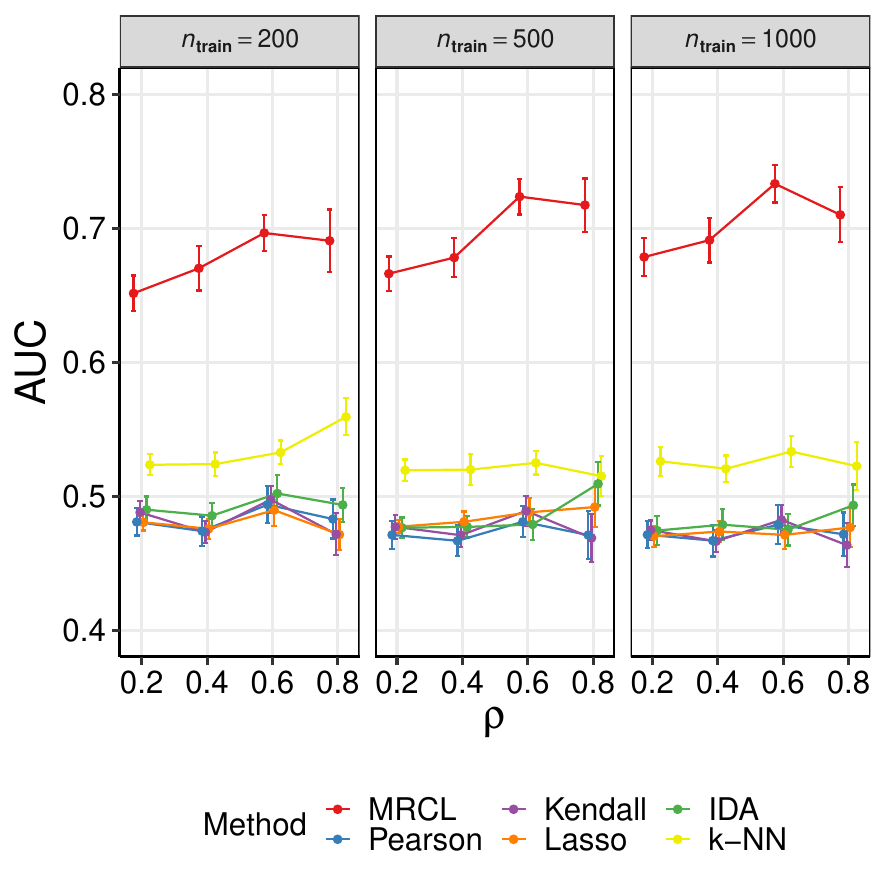}
\caption{Results for data set D1 (yeast data), random sampling. Area under the ROC curve (AUC; with respect to causal relationships
determined from unseen interventional data), as a function of the fraction $\rho$ of labels available (labels were sampled at random). Results are shown for three training data sample sizes $n_\mathrm{train}$.
Results are mean values over 25 iterations and error bars indicate standard error of the mean. Additional results for the PC algorithm appear in Appendix~\ref{sec:app-figs} (see text for details).
}
\label{YEAST_GLOBAL}
\end{figure}

Figure~\ref{YEAST_GLOBAL} shows the area under the ROC curve (AUC) as a function of the proportion $\rho$ of entries in $A(C)$ that were observed, for the three sample sizes.
Results were averaged over 25 iterations.
MRCL showed good performance relative to the other approaches for all 12 considered combinations of $n_{\text{train}}$ and $\rho$
(for the other methods shown in Figure~\ref{YEAST_GLOBAL}, any variation in performance with $\rho$ was solely due to the changing test set as these methods do not use the background knowledge $\Phi$). 
Results for PC, which provides a point estimate of a graphical object, are shown as points on the ROC plane 
for the 12 different regimes in Appendix~\ref{sec:app-figs} (Fig.~\ref{fig:yeast_roc_entry}). 
We considered also the transitive closure (motivated by the nature of the experimental data) and 
exploiting the background information $\Phi$ via additional constraints. 
MRCL performs well relative to the other methods in all regimes (see also the Discussion).

\begin{figure}[t]
\centering
\includegraphics[width = 0.6\textwidth]{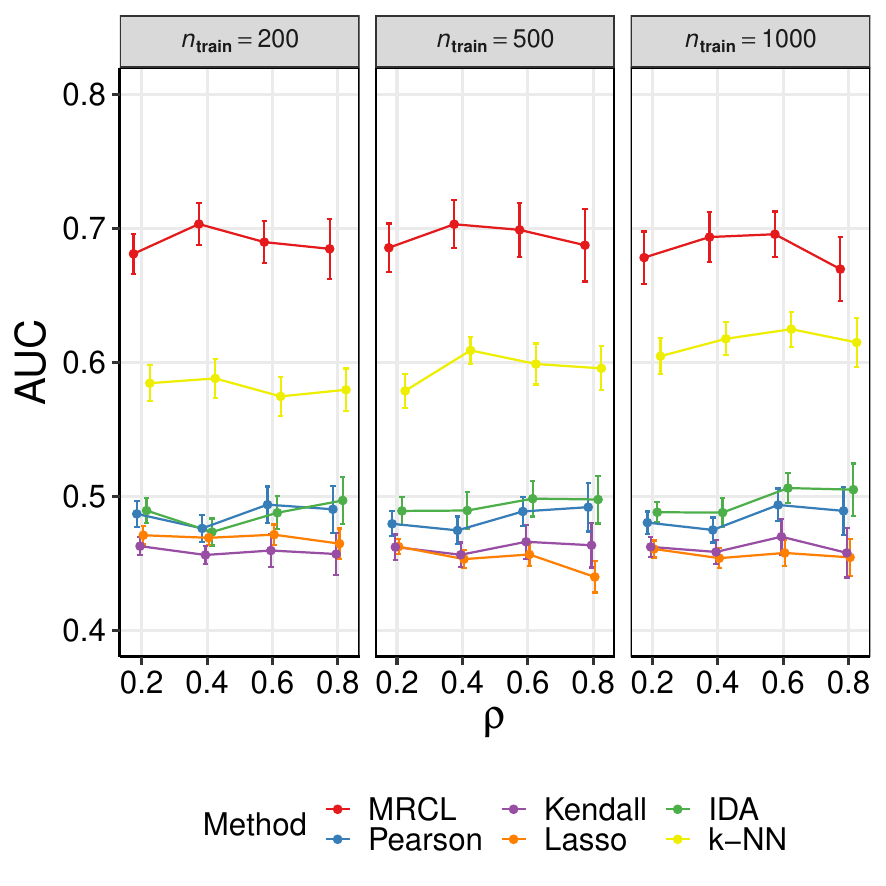}
\vspace{-0.2cm}
\caption{Results for data set D1 (yeast data), row-wise sampling.
As Figure~\ref{YEAST_GLOBAL}, except the subset of labels available to the learner were obtained by sampling entire rows of the causal adjacency matrix. As before, a proportion $\rho$ were sampled. The remaining rows were then used as test data. Additional results for the PC algorithm appear in Appendix~\ref{sec:app-figs} (see text for details).
}
\label{YEAST_ROW}
\end{figure}

In the above results the pairs whose causal relationship was to be predicted were chosen at random (i.e., the set of unlabelled pairs was a random subset of the set of all pairs). In contrast, in some settings it may be relevant to predict the effect of intervening on variable $i$, without knowing the effect of intervening on $i$ on {\it any} other variable. 
For this setting, the unlabelled set should comprise entire rows of the causal adjacency matrix $A(C)$. 
Figure~\ref{YEAST_ROW} considers this case.
To ensure a sufficient number of rows were non-empty, we imposed the additional restriction on the gene subset $C$ 
that at least half of the rows had at least one causal effect.
Results for PC are shown in Appendix~\ref{sec:app-figs} 
(Fig.~\ref{fig:yeast_roc_row})
 as points on the ROC plane. 
As for the random sampling case above, MRCL offers an improvement over the other methods.
k-NN also performs well relative to the other approaches here.

\begin{figure}[p]
\centering
\includegraphics[width = 0.6\textwidth]{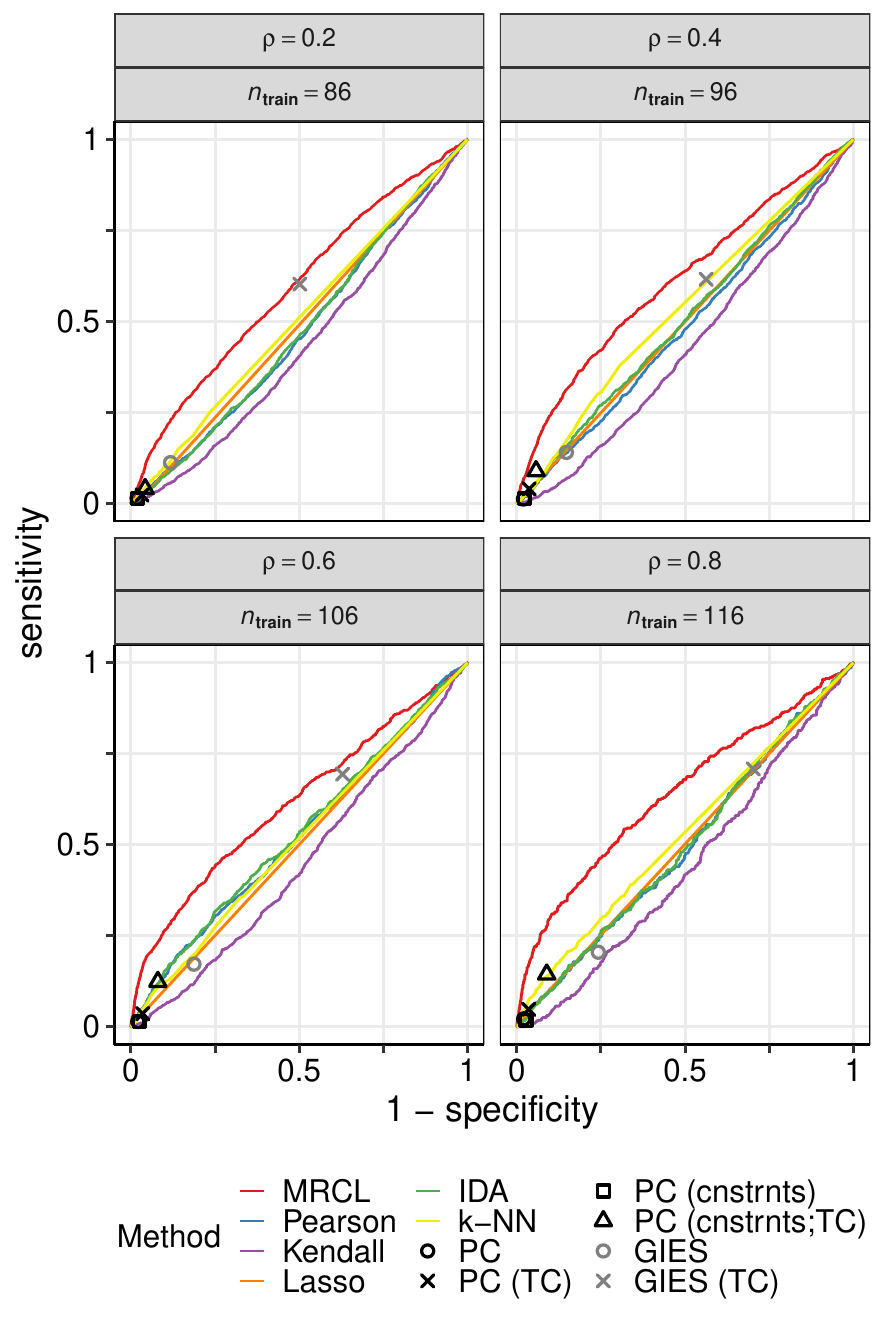}
\caption{Results for data set D1 (yeast data), comparison including GIES, row-wise sampling. 
ROC curves are shown with respect to causal relationships
determined from unseen interventional data.
``TC" indicates use of a transitive closure operation and 
``cnstrnts" indicates that the background information $\Phi$ was included via input constraints.
Results for PC and GIES are shown as points on the ROC plane. 
Note that due to the nature of input required by GIES the data matrices in this example differ from the row-wise sampling example in Figure~\ref{YEAST_ROW}  (see text for details). 
Results are averages over 25 iterations.
}
\label{fig:yeast_roc_gies}
\end{figure}

We additionally compared MRCL with GIES. GIES  and MRCL differ in terms of their required inputs:
In addition to data $\mathcal{D}$, MRCL requires  binary labels on causal relationships via background information $\Phi$, while GIES requires the interventional data itself and metadata specifying the intervention targets.
For row-wise sampling, to allow for a reasonable comparison, we ran GIES providing the interventional data corresponding to the  rows whose labels are provided to MRCL. The same data was also provided as input to the other approaches, including in data set $\mathcal{D}$ for MRCL. This means the data matrices differ from those above, with sample size dependent  on $\rho$, and for MRCL,  $\mathcal{D}$ now includes data that was used to obtain background information $\Phi$ (train/test validity is preserved since it remains the case that all testing is done with respect to  entirely unseen interventions). 
Results appear  in Figure~\ref{fig:yeast_roc_gies},  with PC and GIES shown as a points on the ROC plane.
MRCL  appears to offer an improvement relative to  the other methods (see also the Discussion). 
Note that GIES is not directly applicable to the random sampling setting above since 
it requires the interventional data with respect to all other variables (and not just a subset thereof).

\subsection{Data Set D2: Protein Time-Course Data}

\begin{figure}[t]
\centering
\includegraphics[width = 0.6\textwidth]{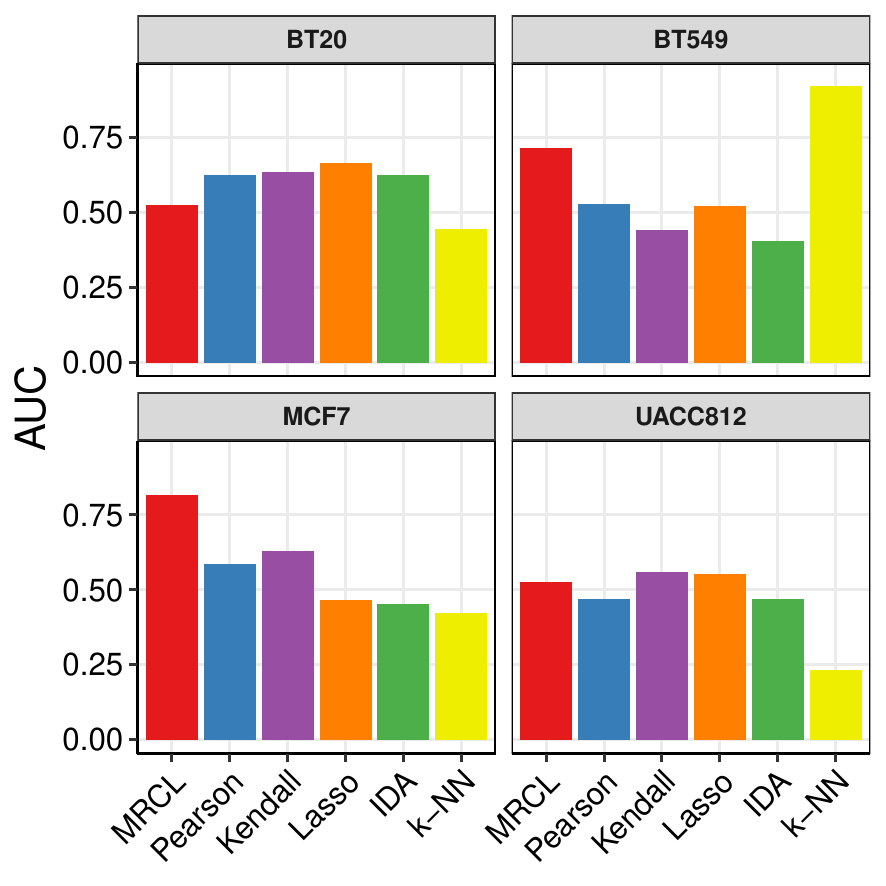}
\caption{Results for data set D2 (protein time course data). Each panel is a different cell line, with its own training and (interventional) test data. AUC is with respect to an entirely held-out intervention. See text for details.}
\label{DREAM}
\end{figure}

\noindent
{\bf Data.} The data consisted of protein measurements for $p=35$ proteins measured at seven time points in four different `cell lines' (BT20, BT549, MCF7 and UACC812; these are laboratory models of human cancer) and under eight growth conditions. The proteins under study act as kinases (i.e., catalysts for a biochemical process known as phosphorylation) and interventions were carried out using kinase inhibitors that block the kinase activity of specific proteins. A total of four intervention regimes were considered, plus a control regime with no interventions.
The data used here were a subset of the complete data set reported in detail in \citet{hill2017} and were also previously used in a Dialogue for Reverse Engineering Assessments and Methods (DREAM) challenge on learning causal networks \citep{hill2016}.

\medskip
\noindent
{\bf Problem set-up.} 
Treating each cell line as a separate, independent problem, the intervention regimes were used to define an interventional `gold standard', in a similar vein as for data set {\bf D1}. This followed the procedure described in detail in \citep{hill2016} with an additional step of taking a majority vote across growth conditions to give a causal gold standard for each cell line $c$.
For each cell line $c$, we formed a data matrix $\mathbf{Z}_c$ consisting of all available data for the $p=35$ proteins except for one of the intervention  regimes. 
The intervention regime not included was a kinase inhibitor targeting the protein mTOR. 
This intervention was entirely held out and used to provide the test labels.
As background knowledge $\Phi_c$ we took as training labels 
causal effects under the other interventions. 
With this set-up, 
the task was to determine the (ancestral) causal effects of the entirely unseen intervention.
Note that each cell line $c$ was treated as an entirely different data set and task, with its own data matrix, background knowledge and interventional test data.

\medskip
\noindent
{\bf Results.} 
Figure~\ref{DREAM} shows AUCs (with respect to changes seen under the test intervention) for each of the four cell lines and each of the methods.
There was no single method that outperformed all others across all four cell lines.
MRCL performed particularly well relative to the other methods for cell lines BT549 and MCF7 (k-NN also performed well for BT549), was competitive for cell line UACC812, but performed less well for cell line BT20.
We note also that, for cell lines BT549 and MCF7, the performance of MRCL was competitive with the best performers in the DREAM challenge and with an analysis reported in \cite{hill2017}. The latter involved a Bayesian model specifically designed for such data. In contrast, MRCL was applied directly to a data matrix comprising all training samples simply collected together.

\subsection{Data Set D3: Human Cancer Data}

\noindent
{\bf Data.} The data consisted of protein measurements for $p=35$ proteins measured 
in $n=820$ human breast cancer samples (from biopsies). The data originate from The Cancer Genome Atlas (TCGA) Project, are described in \cite{akbani2014} and were retrieved from The Cancer Proteome Atlas (TCPA) data portal \cite[][\url{https://tcpaportal.org}; data release version 4.0; Pan-Can 19 Level 4 data]{Li2013}.
Data for many cancer types are available, but here we focus on a single type (breast cancer) 
to minimize the potential for confounding by cancer type.
It is at present difficult to carry out interventions in biopsy samples of this kind. However, we focused on the same 35 proteins as in data set {\bf D2}, whose mutual causal relationships are relatively well-understood, and used a reference causal graph for these proteins based on the biochemical literature \cite[as reported in][]{hill2017}. 

\medskip
\noindent
{\bf Problem set-up.} 
We formed a data set $\mathcal{D}$ consisting of measurements for the $p=35$ proteins for three different sample sizes: (i) $n_{\mathrm{train}}=200$, (ii) $n_{\mathrm{train}}=500$ or (iii) all $n_{\mathrm{train}}=820$ patient samples. For (i) and (ii) patient samples were selected at random.
We then used a random fraction $\rho$ of the reference graph as background knowledge, testing output 
on the (unseen) remainder. 

\begin{figure}[tbh]
\centering
\includegraphics[width = 0.6\textwidth]{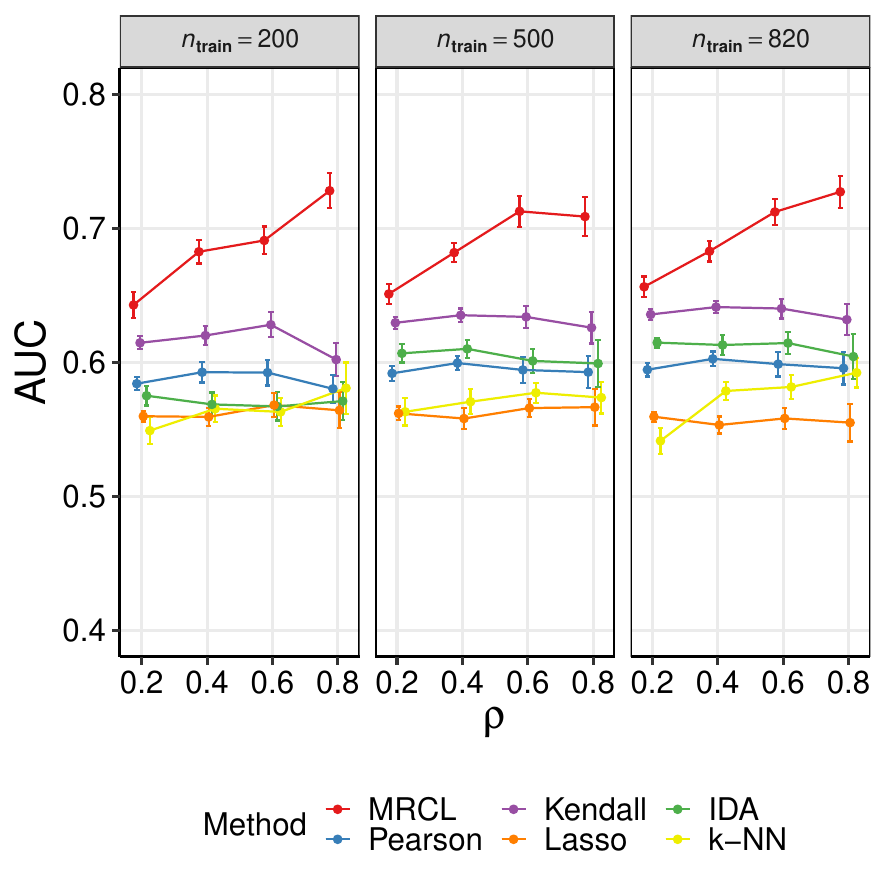}
\vspace{-0.7cm}
\caption{Results for data set D3 (human cancer data). Data are protein measurements from breast cancer patient samples from The Cancer Genome Atlas (TCGA). AUC is with respect to a reference graph based on the (causal) biochemical literature. Results are mean values over 25 iterations and error bars indicate standard error of the mean. See text for details. Additional results appear in Appendix~\ref{sec:app-figs}.}
\label{PANCAN}
\end{figure}

\medskip
\noindent
{\bf Results.} 
Figure~\ref{PANCAN} shows AUCs (with respect to the held-out causal labels) as a function of the proportion $\rho$ of 
causal labels that were  observed, for each of the methods and for the three sample sizes.
Results were averaged over 25 iterations.
MRCL performed well relative to the other methods, with performance improving with $\rho$. 
Results were qualitatively similar for the three sample sizes, with increases in AUC for $n_{\mathrm{train}}=820$ and $n_{\mathrm{train}}=500$ relative to $n_{\mathrm{train}}=250$.
Results for 
PC
are shown in Appendix~\ref{sec:app-figs} (Fig.~\ref{fig:TCPA_roc}) as points on the ROC plane.

 \section{Discussion}
 
In this paper, we showed how
a key aspect of 
causal structure learning can be framed as a machine learning task. 
 Although many available approaches, including those based on DAGs and related graphical models, offer a well-studied framework, we think it may be fruitful to revisit some questions in causality using  machine learning tools.

  In our experiments, based on three real data sets, we found that MRCL performed well relative to a range of graphical model-based approaches. 
  However, two points should be noted regarding these comparative results. 
  First, the various methods differ with respect to their required inputs and the nature of their outputs. This means that in some cases specific methods may not be an ideal fit to the context of the specific data/task (as detailed when presenting the empirical results above).
  Second, the biological systems underlying these data sets  are likely to have features 
  (such as causal insufficiency and cycles)
  that violate one or more of the assumptions of some of these methods.  
  That said, we think biological data sets of the kind we focused on here offer perhaps the best opportunity at present to empirically study causal learning methods and that causal learning tasks of the kind addressed here are highly relevant in many applications, in biology and beyond. Hence, we think that pursuing empirical work on such data is valuable both from methodological and applied points of view. 
	As more interventional data become available in the future, it will be important to carry out similar analyses in other contexts, in order to better understand the extent to which our findings generalize to other scientific settings.

An open  question from a theoretical point of view is to understand  conditions on data-generating processes needed to permit a discriminative approach as pursued here and we think this will be an interesting  direction for future  work. One point of view---analogous to that used in practical applications of classification---is to estimate the risk of the learner and thereby report an estimate of (causal) efficacy without having to directly consider requirements on the underlying system. We think this approach is acceptable when some causal information  is available, since one can then empirically  test problem-specific efficacy (as in our examples above). This then gives confidence with respect to generalization to new interventions on the system of interest (but does not address the broader theoretical question).

In our approach information on multiple variable pairs is coupled via the classifier but not by global constraints on the  graph. In the scientific settings we focused on we did not consider further coupling via global constraints but such constraints (e.g. enforcing transitivity) could be relevant in some applications and an interesting direction for further work.
The main  advantage of our approach is that it allows  regularities  in the data to emerge via learning, rather than having to be encoded via an explicit  causal or mechanistic model. It also naturally provides some uncertainty quantification, in the sense of scores that can be used to guide decisions or future experimental work.
The main disadvantage relative to methods rooted in DAGs and related graphical models is the lack of a full causal model. Albeit under relatively strong assumptions, DAG-based models, once estimated, can be used to shed light on a huge range of questions concerning causal relationships, including direct and ancestral effects, and details of post-intervention distributions. 
In contrast, our approach in itself provides only estimates of binary causal relationships.
That said, given the efficacy and simplicity of our approach, we think it would be fruitful to consider coupling it to established causal tools in a two-step approach, with our methods used to learn an edge structure in a data-driven manner and this structure used to inform a full analysis in a second step. Such an approach would require some care to avoid bias, and sample splitting techniques that have been studied in high-dimensional statistics could be relevant \citep{wasserman2009,stadler2017}. 

\section{Code Availability}
All computational analysis was performed in R \citep{R2018}.
Source code for MRCL and scripts to generate the empirical results presented in Section~\ref{sec: experimental results} are available at \url{https://github.com/Steven-M-Hill/MRCL}.

\bigskip

\noindent
{\bf Acknowledgements:}
The authors are grateful to the Editor and Reviewers for their constructive feedback on an earlier version of the manuscript. The authors are grateful to Umberto No\`{e} for input on the empirical work and to Oliver Crook for feedback on an earlier version of the manuscript. 
This work was supported by the UK Medical Research Council (University Unit Programme number MC{\textunderscore}UU{\textunderscore}00002/2).
CJO was supported by the ARC Centre of Excellence for Mathematics and Statistics, Australia, and the Lloyd's Register Foundation programme on data-centric engineering at the Alan Turing Institute, UK.
The results shown here are in part based upon data generated by the TCGA Research Network: \url{https://www.cancer.gov/tcga}.

\smallskip
\noindent
{\bf Author contributions:} CJO, DB and SM developed methodology based on original ideas by SM. SMH performed the computational work and CJO contributed theory. CJO and SM wrote the paper with input from SMH and DB.

\bigskip

\appendix
\section{Proof of Results in the Main Text}

In this appendix we provide proofs for the theoretical results in the main text.

\vspace{10pt}

\begin{proof}[Proposition \ref{prop: pseudometric inherit}]
Given $S,S' \in \mathcal{S}_n$, we obtain from the pseudo-metric properties of $d_{\mathcal{P}}$ each of (i) positivity $d_{\mathcal{S}}(S,S') = d_{\mathcal{P}}(\kappa(S), \kappa(S')) \geq 0$, (ii) pseudo-identity $d_{\mathcal{S}}(S,S) = d_{\mathcal{P}}(\kappa(S) , \kappa(S)) = 0$, (iii) symmetry $d_{\mathcal{S}}(S,S') = d_{\mathcal{P}}(\kappa(S),\kappa(S')) = d_{\mathcal{P}}(\kappa(S'),\kappa(S)) = d_{\mathcal{S}}(S',S)$ and (iv) the triangle inequality $d_{\mathcal{S}}(S,S'') = d_{\mathcal{P}}(\kappa(S),\kappa(S'')) \leq d_{\mathcal{P}}(\kappa(S),\kappa(S')) + d_{\mathcal{P}}(\kappa(S'),\kappa(S'')) = d_{\mathcal{S}}(S,S') + d_{\mathcal{S}}(S',S'')$, for all $S,S',S'' \in \mathcal{S}_n$.

Suppose now that $\kappa$ is injective and $d_{\mathcal{P}}$ is a metric on $\mathcal{P}$.
Suppose now that $\kappa$ is injective and $d_{\mathcal{P}}$ is a metric on $\mathcal{P}$.
Then if it holds that $d_{\mathcal{S}}(S,S') = d_{\mathcal{P}}(\kappa(S),\kappa(S'))$ $= 0$, it follows that $\kappa(S) = \kappa(S')$ which (from assumption on $\kappa$) implies $S = S'$ in $\mathcal{S}_n$.
Thus under these additional assumptions, $d_{\mathcal{S}}$ is a metric on $\mathcal{S}_n$.
\end{proof}

\begin{proof}[Proposition \ref{prop: d metric}]
The non-negativity, symmetry and sub-additivity properties of $d_{\mathcal{P}}$ are clear, so all that remains is to establish that $d_{\mathcal{P}}(\pi,\pi') = 0$ implies $\pi = \pi'$.
From the definition of $\mathcal{P}$, both $\pi$ and $\pi'$ are continuous on $\mathcal{Z}_2$.
The result is then immediate from the fact that, since $\pi$ and $\pi'$ are continuous and $\mathcal{Z}_2$ is compact, then $\int_{\mathcal{Z}_2} |\pi(\mathbf{z}') - \pi'(\mathbf{z}')|^2 \; \mathrm{d}\Lambda_2(\mathbf{z}') = 0$ implies $\pi$ and $\pi'$ must be identical as functions on $\mathcal{Z}_2$.
\end{proof}

\begin{proof}[Proposition \ref{prop: consistent}]
Observe that, using Prop. \ref{prop: d metric} for sub-additivity of the metric $d_{\mathcal{P}}$,
\begin{eqnarray*}
d_{\mathcal{S}}(S^{(n)},\tilde{S}^{(n)}) & = &  d_{\mathcal{P}}(\kappa(S^{(n)}),\kappa(\tilde{S}^{(n)})) \\
& \leq & d_{\mathcal{P}}(\kappa(S^{(n)}),\pi) + d_{\mathcal{P}}(\pi,\tilde{\pi}) + d_{\mathcal{P}}(\tilde{\pi},\kappa(\tilde{S}^{(n)}))  \\
& = & d_{\mathcal{P}}(\pi,\tilde{\pi}) +  \|\pi - \kappa(S^{(n)})\|_{L^2(\Lambda_2)} + \|\tilde{\pi} - \kappa(\tilde{S}^{(n)})\|_{L^2(\Lambda_2)}
\end{eqnarray*}
Since $\kappa$ is consistent we have $\|\pi - \kappa(S^{(n)})\|_2 = o_P(1)$ and $\|\tilde{\pi} - \kappa(\tilde{S}^{(n)})\|_2 = o_P(1)$.
This completes the proof.
\end{proof}

\begin{proof}[Proposition \ref{prop: convergence rate}]
This proof extends the simpler proof given for the univariate case in Theorem 6.11 of \cite{Wassermann2006}.
For convenience, and without loss of generality, we suppose that $\mathcal{Z}_2 = [0,1]^2$.
It will be convenient in this section to re-assign the notation $\mathbf{z}$ as a dummy variable in $\mathcal{Z}_2$ (instead of in $\mathcal{Z}_p$).
Let 
$$
p_{i,j} = \int_{B_{i,j}} \pi \mathrm{d}\Lambda_2
$$
be the probability mass assigned to 
\begin{align*}
B_{i,j} &= \left[z_{\text{min}} + (z_{\text{max}}-z_{\text{min}})\frac{i-1}{M} , z_{\text{min}} + (z_{\text{max}}-z_{\text{min}})\frac{i}{M} \right) \\
& \qquad \times \left[z_{\text{min}} + (z_{\text{max}}-z_{\text{min}})\frac{j-1}{M} , z_{\text{min}} + (z_{\text{max}}-z_{\text{min}})\frac{j}{M} \right) ,
\end{align*}
so that, from binomial properties, the mean and variance of the histogram estimator $\kappa(S^{(n)})(\mathbf{z})$ at the point $\mathbf{z} \in \mathcal{Z}_2$ are
\begin{eqnarray*}
m(\mathbf{z}) & = & \frac{p_{i,j}}{h^2} \\
v(\mathbf{z}) & = & \frac{p_{i,j}(1 - p_{i,j})}{n h^4}.
\end{eqnarray*}
Let $b(\mathbf{z}) = m(\mathbf{z}) - \pi(\mathbf{z})$ denote the bias of the histogram estimator.
The mean square of the error $\pi(\mathbf{z}) - \kappa(S^{(n)})(\mathbf{z})$ at a point $\mathbf{z}' \in \mathcal{Z}_2$ can be bias-variance decomposed:
$$
\mathbb{E} \{ [ \pi(\mathbf{z}) - \kappa(S^{(n)})(\mathbf{z}) ]^2 \} = b(\mathbf{z})^2 + v(\mathbf{z})
$$
The aim is to obtain independent bounds on both the bias and variance terms next.

To bound the bias term, Taylor's theorem gives that, for $\mathbf{z}, \mathbf{z}' \in B_{i,j}$,
\begin{eqnarray}
\pi(\mathbf{z}') = \pi(\mathbf{z}) + (\mathbf{z}' - \mathbf{z})^\top \cdot \nabla \pi(\mathbf{z}) + \frac{1}{2} (\mathbf{z}' - \mathbf{z})^\top \; R_{i,j}(\mathbf{z}) \; (\mathbf{z}' - \mathbf{z}) \label{eqn:Taylor}
\end{eqnarray}
where the remainder term satisfies
\begin{eqnarray*}
\|R_{i,j}(\mathbf{z})\|_{\max} & \leq & \sup_{\mathbf{z}'' \in B_{i,j}} \| \nabla \nabla^\top \pi(\mathbf{z}'') \|_{\max} \quad \text{(Taylor)} \\
& \leq & \sup_{\mathbf{z}'' \in \mathcal{Z}_2} \| \nabla \nabla^\top \pi(\mathbf{z}'') \|_{\max} \; < \; \infty \quad \text{(continuous on compact domain)} .
\end{eqnarray*}
Here $\|M\|_{\max} = \max\{M_{i,j}\}$ and $\nabla \nabla^\top \pi$ denotes the Hessian, which exists since $\pi$ is twice continuously differentiable in $\mathcal{Z}_2$.
Thus for $\mathbf{z} \in B_{i,j}$, integrating (\ref{eqn:Taylor}):
$$
\int_{B_{i,j}} \pi(\mathbf{z}') \mathrm{d}\Lambda_2(\mathbf{z}') = h^2 \pi(\mathbf{z}) + h^2 \left(\frac{h}{2} \left[ \begin{array}{c} 2i-1 \\ 2j-1 \end{array} \right] - \mathbf{z} \right) \cdot \nabla \pi(\mathbf{z}) + E_{i,j}(\mathbf{z}) 
$$
where the new remainder term can be bounded:
\begin{eqnarray}
|E_{i,j}(\mathbf{z})| & = & \left| \frac{1}{2} \int_{B_{i,j}} (\mathbf{z}' - \mathbf{z})^\top \; R_{i,j}(\mathbf{z}) \; (\mathbf{z}' - \mathbf{z}) \mathrm{d}\Lambda_2(\mathbf{z}') \right| \nonumber \\
& \leq & \frac{1}{2} \int_{B_{i,j}} \|\mathbf{z}' - \mathbf{z}\|_2^2 \mathrm{d}\Lambda_2(\mathbf{z}') \times \sup_{\mathbf{z}'' \in \mathcal{Z}_2} \|\nabla \nabla^\top \pi(\mathbf{z}'')\|_{\max} \label{eq:interm eq} \\
& \leq & 8 h^4 \sup_{\mathbf{z}'' \in \mathcal{Z}_2} \| \nabla \nabla^\top \pi(\mathbf{z}'') \|_{\max} \; =: \; C h^4 \nonumber
\end{eqnarray}
where the constant $C$ is independent of $\mathbf{z}$ and $i,j$.
The number 8 (which is not sharp) is obtained from trivial but tedious computation of the integral in (\ref{eq:interm eq}) and bounding each term in the result.
Now, for $\mathbf{z} \in B_{i,j}$, the bias is expressed using (\ref{eqn:Taylor}) as
\begin{eqnarray*}
b(\mathbf{z}) & = & \frac{1}{h^2} \int_{B_{i,j}} \pi(\mathbf{z}') \mathrm{d}\Lambda_2(\mathbf{z}') - \pi(\mathbf{z}) \\
& = & \left(\frac{h}{2} \left[ \begin{array}{c} 2i-1 \\ 2j-1 \end{array} \right] - \mathbf{z} \right) \cdot \nabla \pi(\mathbf{z}) + \frac{1}{h^2} E_{i,j}(\mathbf{z}) .
\end{eqnarray*}
Now we integrate this expression over $\mathbf{x} \in B_{i,j}$:
\begin{eqnarray*}
\int_{B_{i,j}} b^2 \mathrm{d}\Lambda_2 & = & \int_{B_{i,j}} \left\{ \left(\frac{h}{2} \left[ \begin{array}{c} 2i-1 \\ 2j-1 \end{array} \right] - \mathbf{z} \right) \cdot \nabla \pi(\mathbf{z}) + \frac{1}{h^2} E_{i,j}(\mathbf{z}) \right\}^2 \mathrm{d}\Lambda_2(\mathbf{z}) \\
& \leq & \int_{B_{i,j}} \left\{ \left(\frac{h}{2} \left[ \begin{array}{c} 2i-1 \\ 2j-1 \end{array} \right] - \mathbf{z} \right) \cdot \nabla \pi(\mathbf{z}) \right\}^2 \mathrm{d}\Lambda_2(\mathbf{z}) \\
& & + 2 \int_{B_{i,j}} \left| \left(\frac{h}{2} \left[ \begin{array}{c} 2i-1 \\ 2j-1 \end{array} \right] - \mathbf{z} \right) \cdot \nabla \pi(\mathbf{z})  \right| \frac{1}{h^2} |E_{i,j}(\mathbf{z})| \mathrm{d}\Lambda_2(\mathbf{z}) \\
& & + \int_{B_{i,j}} \frac{1}{h^4} E_{i,j}(\mathbf{z})^2 \mathrm{d} \Lambda_2(\mathbf{z}) \\
& \leq & \int_{B_{i,j}} \left\{ \left(\frac{h}{2} \left[ \begin{array}{c} 2i-1 \\ 2j-1 \end{array} \right] - \mathbf{z} \right) \cdot \nabla \pi(\mathbf{z}) \right\}^2 \mathrm{d}\Lambda_2(\mathbf{z}) \\
& & + 2 C h^2 \int_{B_{i,j}} \left| \left(\frac{h}{2} \left[ \begin{array}{c} 2i-1 \\ 2j-1 \end{array} \right] - \mathbf{z} \right) \cdot \nabla \pi(\mathbf{z})  \right| \mathrm{d}\Lambda_2(\mathbf{z}) + C^2 h^2
\end{eqnarray*}
To bound these integrals we use Cauchy-Schwarz:
\begin{eqnarray}
\int_{B_{i,j}} \left\{ \left(\frac{h}{2} \left[ \begin{array}{c} 2i-1 \\ 2j-1 \end{array} \right] - \mathbf{z} \right) \cdot \nabla \pi(\mathbf{z}) \right\}^2 \mathrm{d}\Lambda_2(\mathbf{z}) & \leq & \int_{B_{i,j}} \left\| \frac{h}{2} \left[ \begin{array}{c} 2i-1 \\ 2j-1 \end{array} \right] - \mathbf{z} \right\|_2^2 \| \nabla \pi(\mathbf{z}) \|_2^2 \mathrm{d}\Lambda_2(\mathbf{z}) \nonumber \\
& \leq & \frac{h^2}{2} \int_{B_{i,j}} \| \nabla \pi(\mathbf{z}) \|_2^2 \mathrm{d}\Lambda_2(\mathbf{z})  \label{eqn: pf term 1}
\end{eqnarray}
and
\begin{eqnarray}
\int_{B_{i,j}} \left| \left(\frac{h}{2} \left[ \begin{array}{c} 2i-1 \\ 2j-1 \end{array} \right] - \mathbf{z} \right) \cdot \nabla \pi(\mathbf{z})  \right| \mathrm{d}\Lambda_2(\mathbf{z}) & \leq & \int_{B_{i,j}} \left\| \frac{h}{2} \left[ \begin{array}{c} 2i-1 \\ 2j-1 \end{array} \right] - \mathbf{z} \right\|_2 \| \nabla \pi(\mathbf{z}) \|_2 \mathrm{d}\Lambda_2(\mathbf{z}) \nonumber \\
& \leq & \frac{h}{\sqrt{2}} \int_{B_{i,j}} \| \nabla \pi(\mathbf{z}) \|_2 \mathrm{d}\Lambda_2(\mathbf{z}). \label{eqn: pf term 2}
\end{eqnarray}
Both expressions in (\ref{eqn: pf term 1}) and (\ref{eqn: pf term 2}) are finite since the integrand is continuous and the domain is compact.
The total integrated bias is thus bounded as
\begin{eqnarray*}
\int_{\mathcal{Z}_2} b^2 \mathrm{d}\Lambda_2 & \leq & \frac{h^2}{2} \int_{\mathcal{Z}_2} \| \nabla \pi(\mathbf{z}) \|_2^2 \mathrm{d}\Lambda_2(\mathbf{z}) + C^2 h^2 + O(h^3)
\end{eqnarray*}

To bound the variance term, from the integral form of the mean value theorem we have that, for some $\mathbf{z}_{i,j} \in B_{i,j}$,
$$
p_{i,j} = \int_{B_{i,j}} \pi \mathrm{d}\Lambda_2 = h^2 \pi(\mathbf{z}_{i,j}) .
$$
The application of the integral form of the mean value theorem is valid since $\pi$ is continuous on $\mathcal{Z}_2$.
Then:
\begin{eqnarray*}
\int_{\mathcal{Z}_2} v^2 \mathrm{d}\Lambda_2 & = & \sum_{i,j=1}^M \int_{B_{i,j}} v \mathrm{d}\Lambda_2 \\
& = & \sum_{i,j=1}^M \int_{B_{i,j}} \frac{p_{i,j}(1 - p_{i,j})}{nh^4} \mathrm{d}\Lambda_2 \\
& = & \frac{1}{nh^2} - \frac{1}{nh^2} \sum_{i,j=1}^M p_{i,j}^2 \\
& = & \frac{1}{nh^2} - \frac{h^2}{n} \sum_{i,j=1}^M \pi(\mathbf{z}_{i,j})^2 \\
& = & \frac{1}{nh^2} - \frac{1}{n} \left( \int_{\mathcal{X}_2} \pi^2 \mathrm{d}\Lambda_2 + o(1) \right) \; = \; \frac{1}{nh^2} + O\left( \frac{1}{n} \right)
\end{eqnarray*}
Putting this all together to obtain a bound:
\begin{eqnarray}
\mathbb{E} \|\pi - \kappa(S^{(n)})\|_{L^2(\Lambda_2)}^2 & = & \int_{\mathcal{X}_2} b^2 \mathrm{d}\Lambda_2 + \int_{\mathcal{Z}_2} v \mathrm{d}\Lambda_2 \quad \text{(Fubini)} \nonumber \\
& \leq & \frac{h^2}{2} \int_{\mathcal{Z}_2} \| \nabla \pi(\mathbf{z}) \|_2^2 \mathrm{d}\Lambda_2(\mathbf{z}) + C^2 h^2 + O(h^3) + \frac{1}{nh^2} + O\left( \frac{1}{n} \right) \label{eqn: final bound}
\end{eqnarray}
where $\mathbb{E}$ denotes expectation with respect to sampling of the data $S^{(n)} \sim \Pi$.
From inspection of (\ref{eqn: final bound}), the estimator error vanishes provided that $h$ is chosen such that $n h^2 \rightarrow \infty$. 
Since convergence in expectation implies convergence in probability, we have established that $\|\pi - \kappa(S^{(n)})\|_{L^2(\Lambda_2)} = o_P(1)$. 
The bandwidth $h^*$, which minimizes the upper bound in (\ref{eqn: final bound}), is 
$$
h^* = \frac{1}{n^{1/4}} \left( \frac{2}{\int_{\mathcal{Z}_2} \| \nabla \pi(\mathbf{z}) \|_2^2 \mathrm{d}\Lambda_2(\mathbf{z}) + 2C^2} \right)^{1/4}
$$
and with this choice we have that $\mathbb{E} \|\pi - \kappa(S^{(n)})\|_{L^2(\Lambda_2)}^2 = O_P(n^{-1/2})$.
For $h = h^*$ we have thus established that $\|\pi - \kappa(S^{(n)})\|_{L^2(\Lambda_2)} = O_P(n^{-1/4})$. 
\end{proof}

\section{Consistency of the Classifier} \label{app: convergence appendix}

Let $\mathcal{X}$ be the compact metric space $\mathcal{X} = \bigtimes_{1 \leq i,j \leq M} [0,n]$ from the main text, where $n$ (the number of points in each scatter plot) is fixed.
Let $\mathcal{Y} = \mathbb{R}$, so that $\{-1,+1\} \subset \mathcal{Y}$.
This section studies the performance of the classifier $\hat{c} : \mathcal{X} \rightarrow \{-1,+1\}$, $\hat{c}(\mathbf{x}) = \text{sign}(\hat{f})$, where $\hat{f}$ is the Laplacian-regularized least squares method from \eqref{eq: LRLS} in the main text, trained on labelled data $\{(\mathbf{x}_{[k]},y_{[k]}) : [k] \in \mathcal{L} \}$ and unlabelled data $\{\mathbf{x}_{[k]} : [k] \in \mathcal{U} \}$, where $\mathbf{x}_{[k]} \in \mathcal{X}$ and $y_{[k]} \in \mathcal{Y}$.
To this end, we must establish a context in which the data pairs $(\mathbf{x}_{[k]},y_{[k]})$ can be considered to be generated.
Let $\rho_{\mathcal{X},\mathcal{Y}}$ be a probability distribution on $\mathcal{X} \times \mathcal{Y}$, with marginals $\rho_{\mathcal{X}}, \rho_{\mathcal{Y}}$ and conditional $\rho_{\mathcal{Y}|\mathcal{X}}$.
In this theoretical investigation we suppose that all data are generated independently from $\rho_{\mathcal{X},\mathcal{Y}}$, with the values $\{y_{[k]} : [k] \in \mathcal{U} \}$ being withheld.

For a generic classifier $c : \mathcal{X} \rightarrow \{-1,+1\}$, define the misclassification rate
$$
\mathcal{R}(c) = \frac{1}{2} \int | y - c(\mathbf{x}) | \mathrm{d}\rho_{\mathcal{X},\mathcal{Y}}(\mathbf{x},y) .
$$
This is minimized by $c_\rho(\mathbf{x}) := \text{sign}(f_\rho(\mathbf{x}))$ where $f_\rho : \mathcal{X} \rightarrow \mathcal{Y}$ is the (typically unavailable) regression function
\begin{eqnarray*}
f_\rho(\mathbf{x}) = \int y \; \mathrm{d}\rho_{\mathcal{Y}|\mathcal{X}}(y | \mathbf{x}) . \label{eq: f rho}
\end{eqnarray*}
Thus the quantity $\mathcal{R}(c_\rho)$ captures the intrinsic difficulty of the classification task.
A classifier $\hat{c}$ is said to be \emph{consistent} (either in expectation, with high probability, etc.) if $\mathcal{R}(\hat{c}) \rightarrow \mathcal{R}(c_\rho)$ in the limit $m_{\mathcal{L}} \rightarrow \infty$ of infinite labelled data (with convergence either in expectation, with high probability, etc.).
Our consistency argument is based around the following straight-forward bound:

\begin{lemma}
Fix $\epsilon > 0$ and let $\mathcal{X}_{\epsilon} := \{\mathbf{x} \in \mathcal{X} : |f_\rho(\mathbf{x})| < \epsilon\}$.
Then
$$
\mathcal{R}(\hat{c}) \leq \mathcal{R}(c_\rho) + \rho_{\mathcal{X}}(\mathcal{X}_\epsilon) + \frac{1}{2 \epsilon} \| \hat{f} - f_\rho \|_{ L^1(\rho_{\mathcal{X}}) } ,
$$
where $\rho_{\mathcal{X}}(\mathcal{X}_\epsilon)$ denotes the $\rho_{\mathcal{X}}$-measure of the set $\mathcal{X}_\epsilon$.
\end{lemma}
\begin{proof}
For all $\mathbf{x} \in \mathcal{X}$, $y \in \mathcal{Y}$, we have that
$$
|y - \text{sign}(\hat{f}(\mathbf{x}))| \leq |y - \text{sign}(f_\rho(\mathbf{x}))| + |\text{sign}(f_\rho(\mathbf{x})) - \text{sign}(\hat{f}(\mathbf{x}))|
$$
so in particular
\begin{align} 
\mathcal{R}(\hat{c}) \leq \mathcal{R}(c_\rho) + \frac{1}{2} \| \text{sign}(f_\rho) - \text{sign}(\hat{f}) \|_{ L^1(\rho_{\mathcal{X}}) } . \label{eq: part 1 lem}
\end{align}
Now,
\begin{align*}
\| \text{sign}(f_\rho) - \text{sign}(\hat{f}) \|_{ L^1(\rho_{\mathcal{X}}) } &= \underbrace{ \int_{\mathcal{X}_{\epsilon}} | \text{sign}(f_\rho) - \text{sign}(\hat{f}) | \mathrm{d}\rho_{\mathcal{X}} }_{(*)} + \underbrace{ \int_{\mathcal{X} \setminus \mathcal{X}_{\epsilon}} | \text{sign}(f_\rho) - \text{sign}(\hat{f}) | \mathrm{d}\rho_\mathcal{X} }_{**} .
\end{align*}
To bound $(*)$, we note that the integrand is trivially bounded by 2.
To bound $(**)$, we note that if $|f_\rho(\mathbf{x})| > \epsilon$ then $\text{sign}(f_\rho) \neq \text{sign}(\hat{f})$ implies that $|\hat{f}(\mathbf{x}) - f_\rho(\mathbf{x})| > 2 \epsilon$.
Thus
\begin{align}
(*) + (**) \; \leq \; 2 \rho_{\mathcal{X}}(\mathcal{X}_\epsilon) + \int_{\mathcal{X} \setminus \mathcal{X}_{\epsilon}} \frac{|f_\rho(\mathbf{x}) - \hat{f}(\mathbf{x})|}{\epsilon} \mathrm{d}\rho_{\mathcal{X}}(\mathbf{x}) \; = \; 2 \rho_{\mathcal{X}}(\mathcal{X}_\epsilon) + \frac{1}{\epsilon} \| \hat{f} - f_\rho \|_{ L^1(\rho_{\mathcal{X}}) } \label{eq: part 2 lem}
\end{align}
Combining \eqref{eq: part 1 lem} and \eqref{eq: part 2 lem} completes the proof.
\end{proof}

Next we leverage an existing high-probability consistency result established in the regression (as opposed to classification) context:

\begin{theorem} \label{thm: CC2012}
Suppose $f_\rho$ is non-constant and that $\Sigma_K^{- \frac{\alpha}{2}} f_\rho \in L^2(\rho_{\mathcal{X}})$ for some $\alpha \in (0,1]$.
Let $\theta = \frac{1}{(1 + \alpha)(1+s)}$.
Take $\lambda_1 = m_{\mathcal{U}}^\theta$ and $\lambda_2 = m_{\mathcal{L}}^\theta$.
Then there exists a finite constant $C$ such that for any $\delta \in (0,1)$, and for $m_{\mathcal{L}}, m_{\mathcal{U}}$ sufficiently large, we have with probability at least $1 - 8 \delta$ that
\begin{align}
\| \hat{f} - f_\rho \|_{L^1(\rho_{\mathcal{X}}) } \leq C \log\left( \frac{2}{\delta} \right) m_{\mathcal{L}}^{- \alpha \theta} . \label{eq: cao result}
\end{align}
\end{theorem}
\begin{proof}
This result is an immediate consequence of Theorem 5.6 in \cite{Cao2012}, whose bound on the $L^2(\rho_{\mathcal{X}})$ error clearly also implies a bound on the $L^1(\rho_{\mathcal{X}})$ error.
In addition, since our intention in what follows is limited to establishing consistency of the proposed classification method, as opposed to a detailed convergence rate analysis, we have simplified the presentation by stating a slightly weaker but less-verbose upper bound.
\end{proof}

Note how the ``for $m_{\mathcal{U}}$ sufficiently large'' condition in Theorem \ref{thm: CC2012} will typically be automatically satisfied in our context, where the amount of unlabelled data is $m_{\mathcal{U}} = O(p^2)$.
Thus the content of \eqref{eq: cao result} is control over $\hat{f} - f_\rho$ as the number $m_{\mathcal{L}}$ of labeled data is increased.

\begin{corollary} \label{cor: risk bound}
Under the same assumptions as Theorem~\ref{thm: CC2012}, we have with probability at least $1 - 8 \delta$ that
\begin{align}
\mathcal{R}(\hat{c}) \leq \mathcal{R}(c_\rho) + \rho_{\mathcal{X}}(\mathcal{X}_\epsilon) + \frac{C}{2 \epsilon} \log\left( \frac{2}{\delta} \right) m_{\mathcal{L}}^{- \alpha \theta} . \label{eq: combined bound}
\end{align}
\end{corollary}
Corollary \ref{cor: risk bound} makes explicit how the intrinsic difficulty of the classification task depends on the form of $f_\rho$, and in particular the extent to which $|f_\rho(\mathbf{x})| < \epsilon$ occurs in $\mathcal{X}$.
For typical regression functions $f_\rho$ with simple roots in $\mathcal{X}$, it will hold that $\rho_{\mathcal{X}}(\mathcal{X}_\epsilon) = O(\epsilon)$.
An assumption of this form can therefore be used to complete a high probability consistency argument:

\begin{corollary}[Consistency of the Classifier] \label{cor: risk bound 2}
Suppose that $\rho_{\mathcal{X}}(\mathcal{X}_\epsilon) = O(\epsilon^\gamma)$ for some $\gamma > 0$.
Under the same assumptions as Theorem~\ref{thm: CC2012}, there exists a finite constant $\tilde{C}$ such that, with probability at least $1 - 8 \delta$,
$$
\mathcal{R}(\hat{c}) \leq \mathcal{R}(c_\rho) + \tilde{C} \left( \log\left( \frac{2}{\delta}  \right) \right)^{\frac{\gamma}{1 + \gamma}} m_{\mathcal{L}}^{- \frac{\alpha \theta \gamma}{1 + \gamma}} .
$$
In particular, this establishes that the classifier $\hat{c}$ is (with high probability) consistent.
\end{corollary}
\begin{proof}
From the hypothesis, $\exists B_1$, $\epsilon_1$ such that $\rho_{\mathcal{X}}(\mathcal{X}_\epsilon) \leq B_1 \epsilon^\gamma$ for all $\epsilon < \epsilon_1$.
Thus, for $\epsilon < \epsilon_1$ the difference $\mathcal{R}(\hat{c}) - \mathcal{R}(c_\rho)$ can be bounded via \eqref{eq: combined bound} as
\begin{eqnarray*}
\mathcal{R}(\hat{c}) - \mathcal{R}(c_\rho) & \leq &\rho_{\mathcal{X}}(\mathcal{X}_\epsilon) + \frac{C}{2 \epsilon} \log\left( \frac{2}{\delta} \right) m_{\mathcal{L}}^{- \alpha \theta} \\ 
& \leq & B_1 \epsilon^\gamma + \frac{B_2}{\epsilon} \; =: \; J(\epsilon)
\end{eqnarray*}
where $B_2 =  \frac{C}{2} \log\left( \frac{2}{\delta} \right) m_{\mathcal{L}}^{- \alpha \theta}$.
Differentiating $J$ and setting to zero reveals that $J$ is minimized over $(0,\infty)$ at
$$
\epsilon^* = \left( \frac{B_2}{\gamma B_1} \right)^{\frac{1}{1 + \gamma}},
$$
which satisfies $\epsilon^* < \epsilon_1$ for $m_{\mathcal{L}}$ sufficiently large (recall that $m_{\mathcal{L}}$ being sufficiently large was an assumption of Theorem~\ref{thm: CC2012}).
Thus, for $m_{\mathcal{L}}$ sufficiently large,
$$
\mathcal{R}(\hat{c}) - \mathcal{R}(c_\rho) \; \leq \; J(\epsilon^*) \; = \; \left( \gamma^{- \frac{\gamma}{1 + \gamma}} + \gamma^{\frac{1}{1 + \gamma}} \right) B_1^{\frac{1}{1 + \gamma}} B_2^{\frac{\gamma}{1 + \gamma}}
$$
which, upon substitution for $B_2$, yields the required result with the value for the constant $\tilde{C} = \left( \gamma^{- \frac{\gamma}{1 + \gamma}} + \gamma^{\frac{1}{1 + \gamma}} \right) B_1^{\frac{1}{1 + \gamma}} \left( \frac{C}{2} \right)^{\frac{\gamma}{1 + \gamma}}$.
\end{proof}

\newpage
\section{Additional Figures}\label{sec:app-figs}

\begin{figure}[!h]
\centering
\includegraphics[width = 0.67\textwidth]{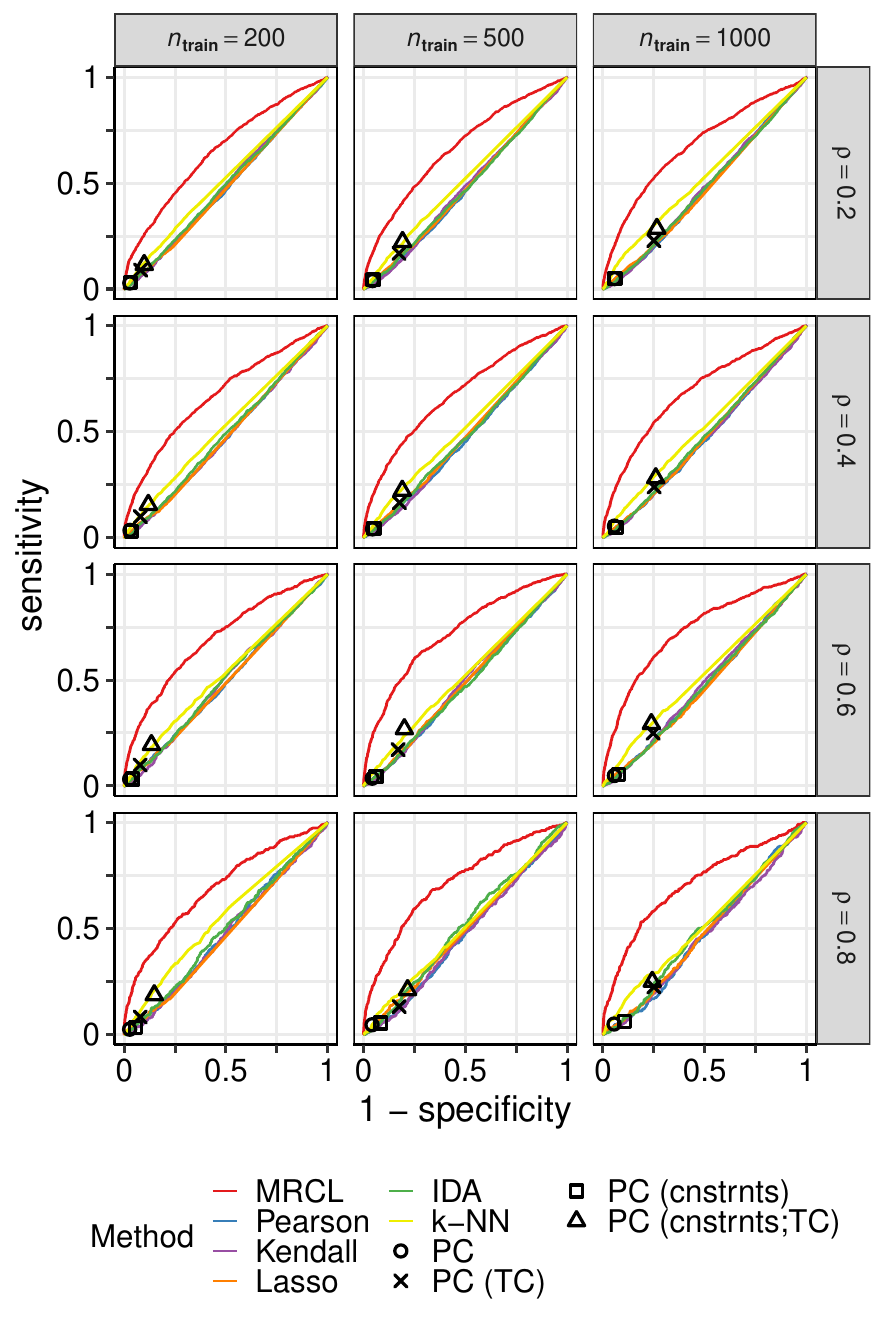}
\caption{Data set D1 (yeast data), random sampling. 
ROC curves are shown with respect to causal relationships
determined from unseen interventional data (see Main Text for details).
Results for PC (which returns a point estimate) are shown as locations on the ROC plane.
``TC" indicates use of a transitive closure operation and 
``cnstrnts" indicates that the background information $\Phi$ was included via input constraints.
[Results shown are for significance level $\alpha =  0.01$ and for a lenient interpretation where possible edges are included. 
Results are averages over 25 iterations.]
}
\label{fig:yeast_roc_entry}
\end{figure}

\begin{figure}
\centering
\includegraphics[width = 0.67\textwidth]{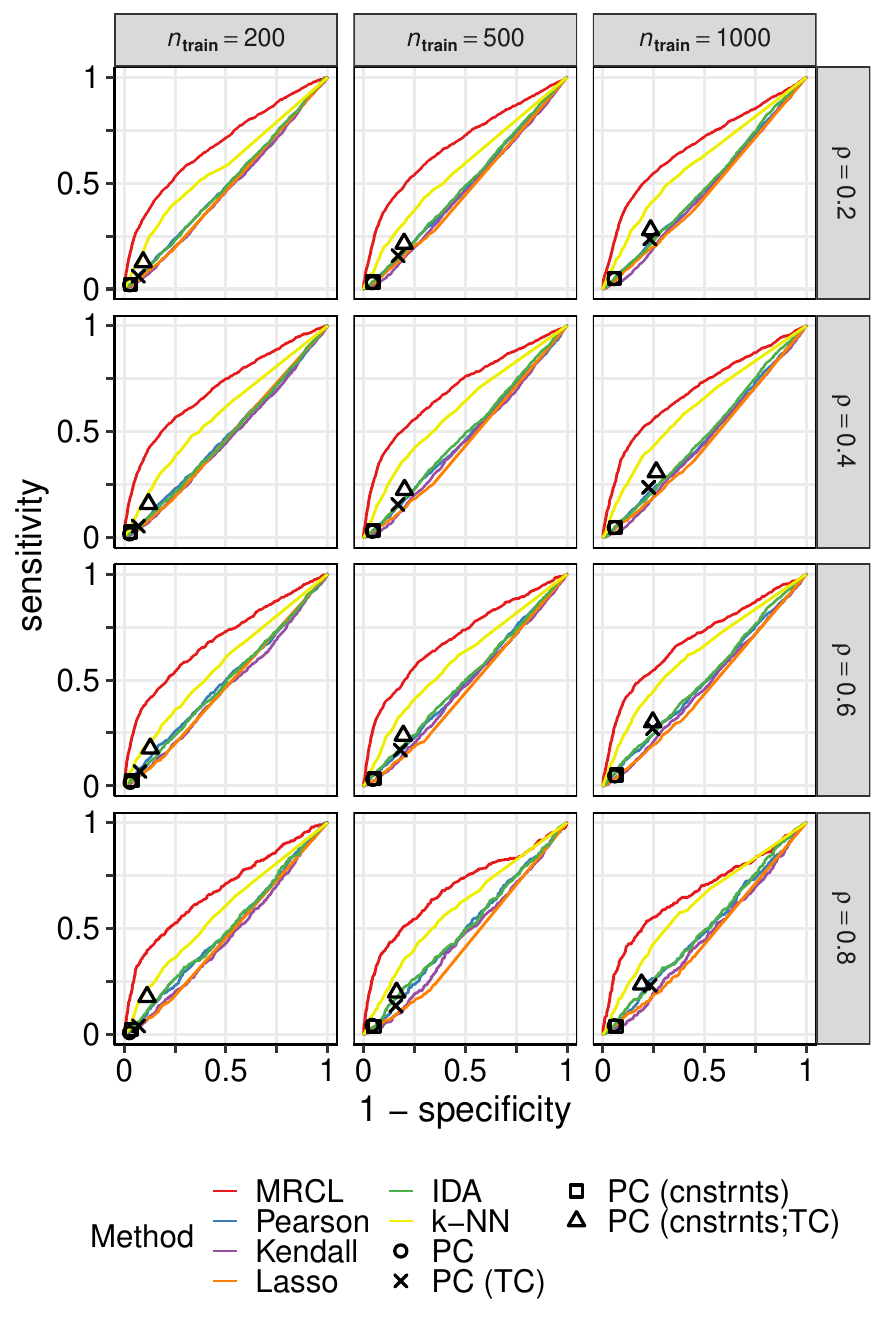}
\caption{Data set D1 (yeast data), row-wise sampling. 
ROC curves are shown with respect to causal relationships
determined from unseen interventional data (see Main Text for details).
Results for PC (which returns a point estimate) are shown as locations on the ROC plane.
``TC" indicates use of a transitive closure operation and 
``cnstrnts" indicates that the background information $\Phi$ was included via input constraints.
[Results shown are for significance level $\alpha =  0.01$ and for a lenient interpretation where possible edges are included.  
Results are averages over 25 iterations.]
}
\label{fig:yeast_roc_row}
\end{figure}

\begin{figure}
\centering
\includegraphics[width = 0.67\textwidth]{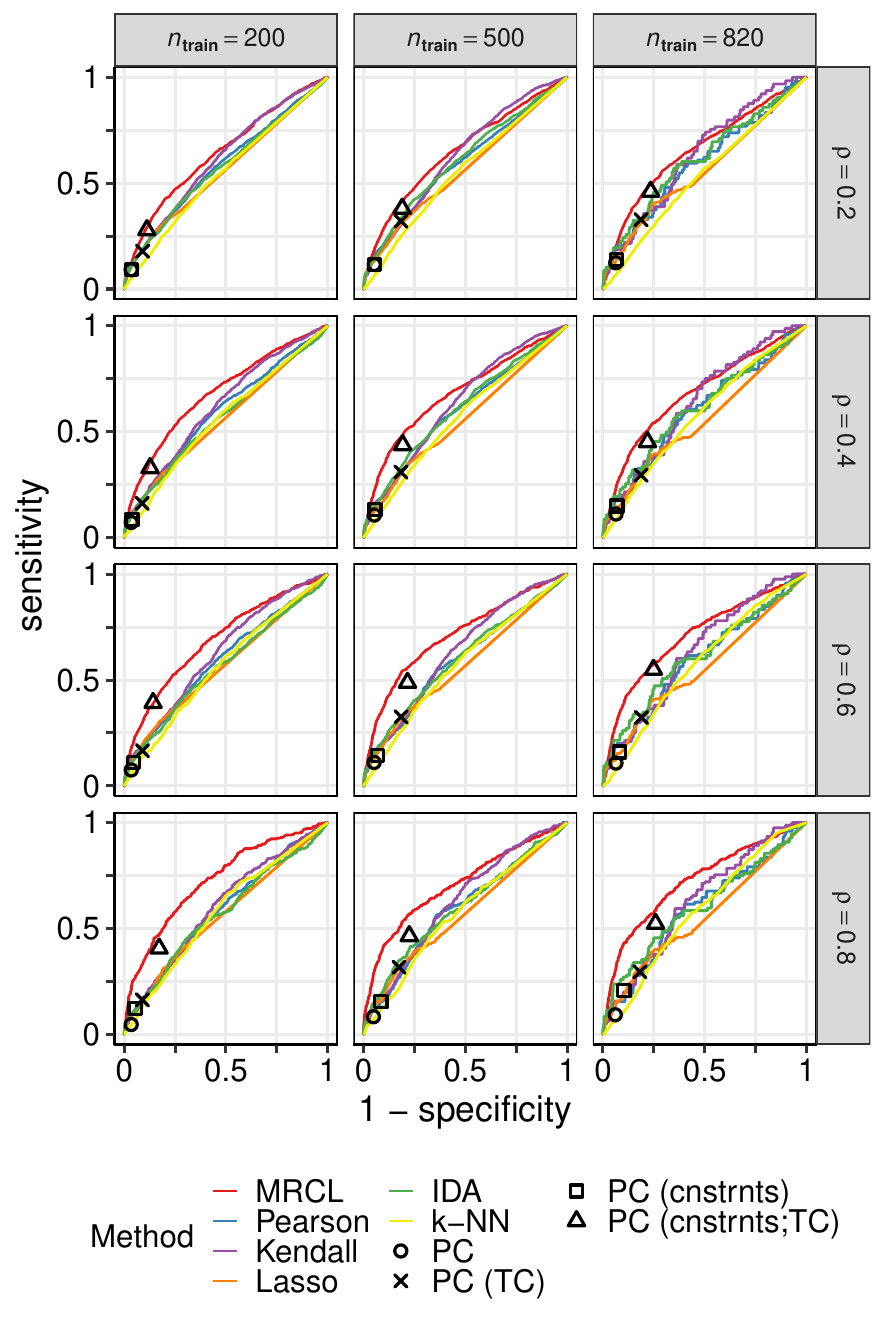}
\caption{Data set D3 (cancer protein data). 
ROC curves are shown with respect to a reference graph determined from the scientific literature (see Main Text for details).
Results for PC (which returns a point estimate) are shown as locations on the ROC plane. 
``TC" indicates use of a transitive closure operation and ``cnstrnts" indicates that the background information $\Phi$ was included via input constraints.
The ``TC'' results are included here for completeness, but we note that the reference graph here encodes direct, rather than ancestral, relationships.
[Results shown are for significance level $\alpha =  0.01$ and for a lenient interpretation where possible edges are included. 
Results are averages over 25 iterations.]
}
\label{fig:TCPA_roc}
\end{figure}


\begin{thebibliography}{31}
\providecommand{\natexlab}[1]{#1}
\providecommand{\url}[1]{\texttt{#1}}
\expandafter\ifx\csname urlstyle\endcsname\relax
  \providecommand{\doi}[1]{doi: #1}\else
  \providecommand{\doi}{doi: \begingroup \urlstyle{rm}\Url}\fi

\bibitem[Akbani et~al.(2014)Akbani, Ng, Werner, Shahmoradgoli, Zhang, Ju, Liu,
  Yang, Yoshihara, Li, Ling, Seviour, Ram, Minna, Diao, Tong, Heymach, Hill,
  Dondelinger, St\"{a}dler, Byers, {Meric-Bernstam}, Weinstein, Broom, Verhaak,
  Liang, Mukherjee, Lu, and Mills]{akbani2014}
R.~Akbani, P.K.S. Ng, H.M. Werner, M.~Shahmoradgoli, F.~Zhang, Z.~Ju, W.~Liu,
  J.Y. Yang, K.~Yoshihara, J.~Li, S.~Ling, E.G. Seviour, P.T. Ram, J.D. Minna,
  L.~Diao, P.~Tong, J.V. Heymach, S.M. Hill, F.~Dondelinger, N.~St\"{a}dler,
  L.A. Byers, F.~{Meric-Bernstam}, J.N. Weinstein, B.M. Broom, R.G.W. Verhaak,
  H.~Liang, S.~Mukherjee, Y.~Lu, and G.B. Mills.
\newblock {A pan-cancer proteomic perspective on The Cancer Genome Atlas}.
\newblock \emph{Nature Communications}, 5:\penalty0 3887, 2014.

\bibitem[Belkin and Niyogi(2008)]{Belkin2008}
M.~Belkin and P.~Niyogi.
\newblock {Towards a theoretical foundation for Laplacian-based manifold
  methods}.
\newblock \emph{Journal of Computer and System Sciences}, 74\penalty0
  (8):\penalty0 1289--1308, 2008.

\bibitem[Belkin et~al.(2006)Belkin, Niyogi, and Sindhwani]{Belkin2006}
M.~Belkin, P.~Niyogi, and V.~Sindhwani.
\newblock Manifold regularization: A geometric framework for learning from
  labeled and unlabeled examples.
\newblock \emph{Journal of Machine Learning Research}, 7:\penalty0 2399--2434,
  2006.

\bibitem[Cao and Chen(2012)]{Cao2012}
Y.~Cao and D.~Chen.
\newblock {Generalization errors of Laplacian regularized least squares
  regression}.
\newblock \emph{Science China Mathematics}, 55\penalty0 (9):\penalty0
  1859--1868, 2012.

\bibitem[Colombo et~al.(2012)Colombo, Maathuis, Kalisch, and
  Richardson]{colombo2012}
D.~Colombo, M.~H. Maathuis, M.~Kalisch, and T.~S. Richardson.
\newblock Learning high-dimensional directed acyclic graphs with latent and
  selection variables.
\newblock \emph{The Annals of Statistics}, 40\penalty0 (1):\penalty0 294--321,
  2012.

\bibitem[Cucker and Zhou(2007)]{Cucker2007}
F.~Cucker and D.~X. Zhou.
\newblock \emph{Learning Theory: An Approximation Theory Viewpoint}.
\newblock Cambridge University Press, 2007.

\bibitem[Fergus et~al.(2009)Fergus, Weiss, and Torralba]{Fergus2009}
R.~Fergus, Y.~Weiss, and A.~Torralba.
\newblock Semi-supervised learning in gigantic image collections.
\newblock In \emph{Proceedings of the 23rd Annual Conference on Neural
  Information Processing Systems}, pages 522--530, 2009.

\bibitem[Grigor'yan(2006)]{Grigoryan2006}
A.~Grigor'yan.
\newblock Heat kernels on weighted manifolds and applications.
\newblock \emph{Cont. Math}, 398:\penalty0 93--191, 2006.

\bibitem[Hauser and B{\"u}hlmann(2012)]{hauser2012}
A.~Hauser and P.~B{\"u}hlmann.
\newblock Characterization and greedy learning of interventional markov
  equivalence classes of directed acyclic graphs.
\newblock \emph{Journal of Machine Learning Research}, 13:\penalty0 2409--2464,
  2012.

\bibitem[Hill et~al.(2016)Hill, Heiser, Cokelaer, Unger, Nesser, Carlin, Zhang,
  Sokolov, Paull, Wong, Graim, Bivol, Wang, Zhu, Afsari, Danilova, Favorov,
  Lee, Taylor, Hu, Long, Noren, Bisberg, {The HPN-DREAM Consortium}, Mills,
  Gray, Kellen, Norman, Friend, Qutub, Fertig, Song, Stuart, Spellman, Koeppl,
  Stolovitzky, {Saez-Rodriguez}, and Mukherjee]{hill2016}
S.M. Hill, L.M. Heiser, T.~Cokelaer, M.~Unger, N.K. Nesser, D.E. Carlin,
  Y.~Zhang, A.~Sokolov, E.O. Paull, C.K. Wong, K.~Graim, A.~Bivol, H.~Wang,
  F.~Zhu, B.~Afsari, L.V. Danilova, A.V. Favorov, W.S. Lee, D.~Taylor, C.W. Hu,
  B.L. Long, D.P. Noren, A.J. Bisberg, {The HPN-DREAM Consortium}, G.B. Mills,
  J.W. Gray, M.~Kellen, T.~Norman, S.~Friend, A.A. Qutub, Y.~Fertig, E.J.~Guan,
  M.~Song, J.M. Stuart, P.T. Spellman, H.~Koeppl, G.~Stolovitzky,
  J.~{Saez-Rodriguez}, and S.~Mukherjee.
\newblock Inferring causal molecular networks: empirical assessment through a
  community-based effort.
\newblock \emph{Nature Methods}, 13\penalty0 (4):\penalty0 310--318, 2016.

\bibitem[Hill et~al.(2017)Hill, Nesser, Johnson-Camacho, Jeffress, Johnson,
  Boniface, Spencer, Lu, Heiser, Lawrence, Pande, Korkola, Gray, Mills,
  Mukherjee, and Spellman]{hill2017}
S.M. Hill, N.K. Nesser, K.~Johnson-Camacho, M.~Jeffress, A.~Johnson,
  C.~Boniface, S.E.F. Spencer, Y.~Lu, L.M. Heiser, Y.~Lawrence, N.T. Pande,
  J.E. Korkola, J.W. Gray, G.B. Mills, S.~Mukherjee, and P.T. Spellman.
\newblock Context specificity in causal signaling networks revealed by
  phosphoprotein profiling.
\newblock \emph{Cell Systems}, 4\penalty0 (1):\penalty0 73--83, 2017.

\bibitem[Hyttinen et~al.(2012)Hyttinen, Eberhardt, and Hoyer]{hyttinen2012}
A.~Hyttinen, F.~Eberhardt, and P.~O. Hoyer.
\newblock Learning linear cyclic causal models with latent variables.
\newblock \emph{Journal of Machine Learning Research}, 13:\penalty0 3387--3439,
  2012.

\bibitem[Kemmeren et~al.(2014)Kemmeren, Sameith, van~de Pasch, Benschop,
  Lenstra, Margaritis, O'Duibhir, Apweiler, van Wageningen, Ko, van Heesch,
  Kashani, Ampatziadis-Michailidis, Brok, Brabers, Miles, Bouwmeester, {van
  Hooff}, Bakel, Sluiters, Bakker, Snel, Lijnzaad, {van Leenen}, {Groot
  Koerkamp}, and Holstege]{kemmeren2014}
P.~Kemmeren, K.~Sameith, L.A. van~de Pasch, J.J. Benschop, T.L. Lenstra,
  T.~Margaritis, E.~O'Duibhir, E.~Apweiler, S.~van Wageningen, C.W. Ko, S.~van
  Heesch, M.M. Kashani, G.~Ampatziadis-Michailidis, M.O. Brok, N.A.C.H.
  Brabers, A.J. Miles, D.~Bouwmeester, S.R. {van Hooff}, H.~Bakel, E.~Sluiters,
  L.V. Bakker, B.~Snel, P.~Lijnzaad, D.~{van Leenen}, M.J.A. {Groot Koerkamp},
  and F.C.P. Holstege.
\newblock Large-scale genetic perturbations reveal regulatory networks and an
  abundance of gene-specific repressors.
\newblock \emph{Cell}, 157\penalty0 (3):\penalty0 740--752, 2014.

\bibitem[Li et~al.(2013)Li, Lu, Akbani, Ju, Roebuck, Liu, Yang, Broom, Verhaak,
  Kane, Wakefield, Weinstein, Mills, and Liang]{Li2013}
J.~Li, Y.~Lu, R.~Akbani, Z.~Ju, P.~L. Roebuck, W.~Liu, J-Y. Yang, B.M. Broom,
  R.G.W. Verhaak, D.W. Kane, C.~Wakefield, J.N. Weinstein, G.B. Mills, and
  H.~Liang.
\newblock {TCPA: a resource for cancer functional proteomics data}.
\newblock \emph{Nature Methods}, 10\penalty0 (11):\penalty0 1046--1047, 2013.

\bibitem[Lopez-Paz et~al.(2015)Lopez-Paz, Muandet, Sch{\"o}lkopf, and
  Tolstikhin]{lopez2015}
D.~Lopez-Paz, K.~Muandet, B.~Sch{\"o}lkopf, and I.~Tolstikhin.
\newblock Towards a learning theory of causation.
\newblock In \emph{Proceedings of the 32nd International Conference on Machine
  Learning}, pages 1452--1461, 2015.

\bibitem[Maathuis et~al.(2009)Maathuis, Kalisch, and
  B{\"u}hlmann]{maathuis2009}
M.H. Maathuis, M.~Kalisch, and P.~B{\"u}hlmann.
\newblock Estimating high-dimensional intervention effects from observational
  data.
\newblock \emph{Annals of Statistics}, 37\penalty0 (6A):\penalty0 3133--3164,
  2009.

\bibitem[Maathuis et~al.(2010)Maathuis, Colombo, Kalisch, and
  B{\"u}hlmann]{maathuis2010}
M.H. Maathuis, D.~Colombo, M.~Kalisch, and P.~B{\"u}hlmann.
\newblock Predicting causal effects in large-scale systems from observational
  data.
\newblock \emph{Nature Methods}, 7\penalty0 (4):\penalty0 247--248, 2010.

\bibitem[Meinshausen et~al.(2016)Meinshausen, Hauser, Mooij, Peters, Versteeg,
  and B{\"u}hlmann]{meinshausen2016}
N.~Meinshausen, A.~Hauser, J.M. Mooij, J.~Peters, P.~Versteeg, and
  P.~B{\"u}hlmann.
\newblock Methods for causal inference from gene perturbation experiments and
  validation.
\newblock \emph{Proceedings of the National Academy of Sciences}, 113\penalty0
  (27):\penalty0 7361--7368, 2016.

\bibitem[Mooij et~al.(2016)Mooij, Peters, Janzing, Zscheischler, and
  Sch{{\"o}}lkopf]{mooij2016}
J.M. Mooij, J.~Peters, D.~Janzing, J.~Zscheischler, and B.~Sch{{\"o}}lkopf.
\newblock Distinguishing cause from effect using observational data: methods
  and benchmarks.
\newblock \emph{Journal of Machine Learning Research}, 17\penalty0
  (32):\penalty0 1--102, 2016.

\bibitem[Pearl(2009)]{pearl2009}
J.~Pearl.
\newblock \emph{Causality}.
\newblock Cambridge University Press, 2009.

\bibitem[Peters et~al.(2016)Peters, B{\"u}hlmann, and Meinshausen]{Peters2015}
J.~Peters, P.~B{\"u}hlmann, and N.~Meinshausen.
\newblock Causal inference using invariant prediction: identification and
  confidence intervals.
\newblock \emph{Journal of the Royal Statistical Society: Series B},
  78\penalty0 (5):\penalty0 947--1012, 2016.

\bibitem[{R Core Team}(2018)]{R2018}
{R Core Team}.
\newblock \emph{R: A Language and Environment for Statistical Computing}.
\newblock R Foundation for Statistical Computing, Vienna, Austria, 2018.
\newblock URL \url{https://www.R-project.org/}.

\bibitem[Richardson(1996)]{richardson1996}
T.~Richardson.
\newblock A discovery algorithm for directed cyclic graphs.
\newblock In \emph{{Proceedings of the Twelfth International Conference on
  Uncertainty in Artificial Intelligence}}, pages 454--461, 1996.

\bibitem[Spirtes(1995)]{spirtes1995}
P.~Spirtes.
\newblock Directed cyclic graphical representations of feedback models.
\newblock In \emph{{Proceedings of the Eleventh Conference on Uncertainty in
  Artificial Intelligence}}, pages 491--498, 1995.

\bibitem[Spirtes et~al.(2000)Spirtes, Glymour, and Scheines]{spirtes2000}
P.~Spirtes, C.~N. Glymour, and R.~Scheines.
\newblock \emph{Causation, Prediction, and Search}.
\newblock MIT press, 2000.

\bibitem[St{\"a}dler and Mukherjee(2017)]{stadler2017}
N.~St{\"a}dler and S.~Mukherjee.
\newblock Two-sample testing in high dimensions.
\newblock \emph{Journal of the Royal Statistical Society: Series B},
  79\penalty0 (1):\penalty0 225--246, 2017.

\bibitem[Tibshirani(1996)]{Tibshirani1996}
R.~Tibshirani.
\newblock Regression shrinkage and selection via the lasso.
\newblock \emph{Journal of the Royal Statistical Society: Series B},
  58\penalty0 (1):\penalty0 267--288, 1996.

\bibitem[Wand and Jones(1994)]{wand1994}
M.P. Wand and M.C. Jones.
\newblock \emph{Kernel Smoothing}.
\newblock CRC Press, 1994.

\bibitem[Wang et~al.(2008)Wang, Jebara, and Chang]{Wang2008}
J.~Wang, T.~Jebara, and S.-F. Chang.
\newblock Graph transduction via alternating minimization.
\newblock In \emph{Proceedings of the 25th International Conference on Machine
  Learning}, pages 1144--1151, 2008.

\bibitem[Wasserman and Roeder(2009)]{wasserman2009}
L.~Wasserman and K.~Roeder.
\newblock High dimensional variable selection.
\newblock \emph{Annals of Statistics}, 37\penalty0 (5A):\penalty0 2178, 2009.

\bibitem[Wassermann(2006)]{Wassermann2006}
L.~Wassermann.
\newblock \emph{{All of Nonparametric Statistics}}.
\newblock Springer, 2006.

\end{thebibliography}
\end{document}